# DCENWCNet: A Deep CNN Ensemble Network for White Blood Cell Classification with LIME-Based Explainability

Sibasish Dhibar[a*], Preeti Preeti[c]
[a]School of Artificial Intelligence,
[a]Amrita Vishwa Vidyapeetham, Bengaluru, 502285, India
[c]Baylor College of Medicine, Houston, USA
d_sibasish@blr.amrita.edu[a], preeti@ma.iitr.ac.in[c]

**Abstract:**

White blood cells (WBC) are key parts of our immune system, and they protect our body against infections by eliminating viruses, bacteria, parasites and fungi. The number of WBC class and the total number of WBCs provide key information about our health status. A traditional technique, convolutional neural networks (CNN), a deep learning (DL) architecture, can classify the blood cell from a part of an object and perform object recognition. Various CNN models exhibit potential; however, their development often involves ad-hoc processes that neglect unnecessary layers, leading to issues with unbalanced datasets and insufficient data augmentation. To address these challenges, we propose a novel ensemble approach that integrates three CNN architectures, each uniquely configured with different dropout and max-pooling layer settings to enhance feature learning. This ensemble model, named DCENWCNet, effectively balances the bias-variance trade-off. When evaluated on the widely recognized Rabbin-WBC dataset, our model outperforms existing state-of-the-art networks, achieving highest mean accuracy. Additionally, it demonstrates superior performance in precision, recall, F1-score, and Area Under the ROC Curve (AUC) across all categories. To investigate deeper into the interpretability of classifiers, we apply reliable post-hoc explanation techniques local interpretable model-agnostic explanations (LIME). These techniques approximate the behavior of a black-box model by revealing the relationships between feature values and predictions. Interpretable outcomes enable users to comprehend and validate the model's predictions, thereby increasing their confidence in the automated diagnosis.

**Keywords:** White blood cell, deep learning, convolution neural network, ensemble network, medical image analysis

## 1. Introduction

Blood comprises numerous parts, containing sugars, hormones, proteins, minerals, carbon dioxide, and blood cells, which are categorized into three major types: red blood cells (RBCs), WBCs, and platelets. These cells, distinct in color, shape, and structure, fulfill specific bodily roles. For illustration, while RBCs transport oxygen throughout the body, WBCs defend against infectious diseases and foreign bodies, and platelets ensure blood clotting, which is vital for stopping bleeding. The various types of WBCs, namely monocytes, eosinophils, basophils, lymphocytes, and neutrophils (White J., 2020), are instrumental in combating various infections (see Fig. 1). The differential count of the WBCs is done in a complete blood cell count (CBC), which is employed to detect any abnormality. Any variation from the average size, maturity, or number of healthy blood cells is analyzed for diagnosing various hematological diseases.

[*] Corresponding author: d_sibasish@blr.amrita.edu

An increased number of WBCs often hints at an infection. For instance, in CBC, an escalation in the number of immature leukocytes can be linked with leukemia. An excess or lack of WBCs may cause various diseases (Ullah et al., 2016). Diagnoses of these diseases are carried out by blood tests. These blood tests are also performed in order to monitor the results of chemotherapy and radiation therapy. WBCs are extensively distributed throughout human blood, lymph, and various tissues, playing a key role in maintaining the body's immune function. Nonetheless, their detection and classification remain challenging owing to the human body's RBC-to-WBC and complexity involved in accurately identifying the classes of WBCs in a blood smear sample. Recently, there have been significant advances in deep learning techniques applied to medicine, underlining their potential in detecting and classifying WBCs in blood samples.

Different blood cell classification algorithms are primarily focused on classifying WBCs. These algorithms leverage features, appearances, textures, patterns, and several attributes of WBCs extracted using feature engineering techniques in deep learning, which play a crucial role in image classification. Currently, image processing techniques are devoted for classification, segmentation, feature extraction, and identification of WBCs (Krizhevsky et al., 2017). Various deep learning (DL) models have been introduced to classify WBC images through both manual processes and computer-aided diagnostic machine learning algorithms based on image processing techniques like k-means clustering, decision trees, support vector machines (SVM), and the integration of CNN and RNN, as well as pre-trained models, which are reported in the research paper Sánchez (2017). These customized CNN models can be further enhanced to achieve higher accuracy rates and improved statistical results.

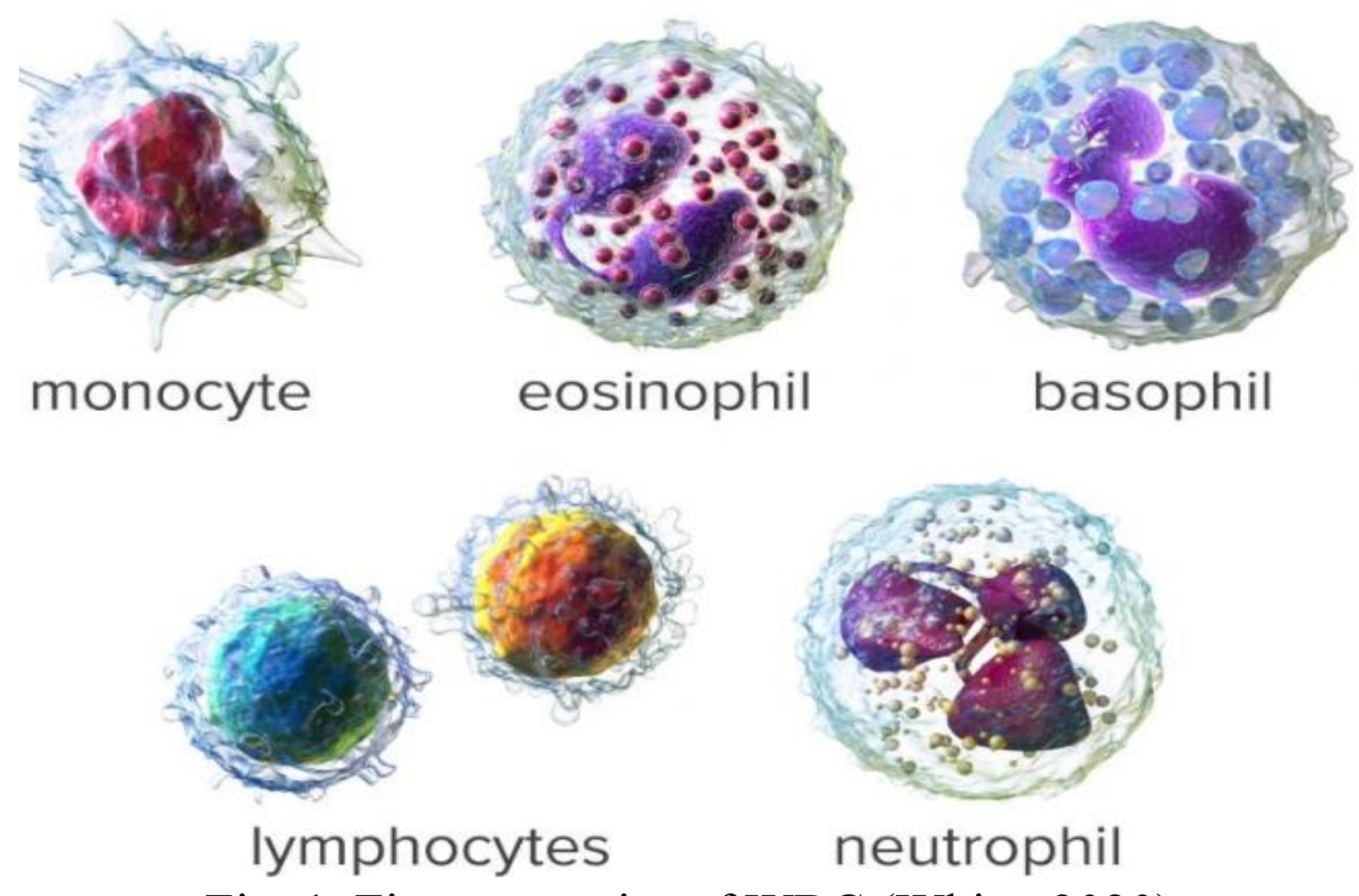

Fig. 1. Five categories of WBC (White, 2020)

### 1.1 Motivation of the research work

The type and count of WBCs are crucial for diagnosing and managing several diseases. CNN based methods have been identified as effective solutions for WBC classification (Habibzadeh et al., 2013; Sharma et al., 2019; Wang et al., 2018). While CNN-based methodologies ensure high classification accuracy through an end-to-end learning schema, their performance can be further enriched by using deep features extracted at numerous levels of CNN architecture. To this end, incorporating feature selection techniques can boost WBC classification accuracy by identifying more discriminative features within a lower-dimensional space. Applying feature selection methods to various CNNs can further improve classification accuracy. This research paper proposes and evaluates the combination of feature selection techniques. It objectives to assess individual features versus simple feature combinations and compare the effectiveness of feature selection methods with simple feature combinations to determine the optimal number of selected features.

The aim of this research paper is to improve classification accuracy by employing multiple customized CNNs through tailored dropout rates and maxpooling strategies. The DCNNs are precisely designed with minimized dropout layers to facilitate effective learning at diverse feature levels, complemented by corresponding max-pooling layers. The final prediction is decided by aggregating the highest accuracy from individual DCNN outputs. This DL model independently learns, trains, validates, and refines features extracted from multiple images. This innovative approach aids in extracting features and patches from input images, thereby improving recognition rates, reducing computational time, and mitigating overfitting issues (Yang et al., 2017). Additionally, the trained images from the pre-trained models are fed into a CNN, enhancing classification accuracy. Ultimately, the fine-tuned models categorize WBCs into five distinct groups. To improve the interpretability of the model's predictions, LIME (Korica et al., 2021) is applied.

### 1.2. Layout of research work

The frame of this paper is organized as follows: Section 2 provides a literature review on WBC studies, identifies research gaps, and highlights the contributions of this study. Section 3 describes the preliminaries of deep learning, the proposed model, and the methodology. Section 4 depicts the dataset, preprocessing steps, and software experiments, along with performance metrics and statistical results. Section 5 discusses the results of the proposed research, supported by graphical and tabular data. Finally, Section 6 presents the concluding remarks.

## 2. Related research articles review

In this section, several algorithms for WBC classification are briefly described, particularly those categorized into three classes. Additionally, the LIME explainability of the predicted outcomes is discussed.

The WBC classification generally falls into three primary techniques: microscopy, flow cytometry, and machine learning (ML). Microscopy-based methods offer high accuracy but are hindered by inefficiencies and prolonged cell culture cycles (Nazlibilek et al., 2014). Flow cytometry has become a widely used procedure for blood sample analysis, yet it presents a major limitation it destroys blood samples, making retrospective WBC studies unfeasible (Othman et al., 2017). In contrast, ML has emerged as the most widely adopted approach due to its simplicity, reliability, and robustness in WBC classification (Ravikumar, 2016). Within the ML framework, WBC classification is achieved through three distinct methodologies: traditional ML algorithms (Nassar et al., 2019; Habibzadeh et al., 2013; Huai et al. 2015; Agaian et al., 2018; Al-Dulaimi et al., 2018; Musliman et al., 2021; Cengil et al. 2022; Ahmad et al. 2023; Wang et al. 2024; Zeng et al. 2024); deep learning (DL) (Duan et al. 2019; Liang et al., 2018; Jiang et al., 2018; Ye et al., 2019; Patil et al., 2021; Khouani et al., 2020; S. Sharma et al., 2022; Girdhar et al., 2022; Dong et al., 2023; Amine Tahiri et al., 2023; Yentrapragada, 2023; Ratheesh and Breethi, 2024; Perveen et al. 2024; Chen et al., 2025; Li et al., 2025); and hybrid methods of ML and DL (Zhao et al., 2017; Hegde et al., 2018; Sánchez et al., 2017; Baydilli & Atila, 2020; Özyurt, 2020; Baghel et al., 2022; Ha et al., 2022; Zhu et al., 2023; ssDong et al. 2024; Aksoy, 2024).

Traditional ML techniques have emphasized the weight of morphological features in accurately classifying WBCs (Nassar et al., 2019). For instance, combining shape, intensity, and texture features with a support vector machine (SVM) classifier achieved an 84% accuracy in classifying 140 digital blood smear images across five WBC types (Habibzadeh et al. 2013). Similarly, an algorithm integrating synthetic features with a random forest (RF) classifier attained a 95.4% accuracy in classifying 800 images of five WBC classes (Huai et al. 2015). Additionally, applying bi-spectral invariant features with SVM and classification trees resulted in a 96.13% average accuracy in distinguishing 10 WBC types across three datasets: Cellavision, ALL-IDB, and Wadsworth Center (Al-Dulaimi et al., 2018). Furthermore, applying multiple features-such as texture, spatial, and spectral data to an SVM classifier for hyperspectral images of five WBC classes led to a 98.3% classification accuracy (Duan et al., 2019). Musliman et al. (2021) utilized HSV segmentation and GLCM for WBC identification, comparing KNN, NBC, and MLP techniques, with accuracies of 82%, 80%, and 94% respectively on testing dataset. Cengil et al. (2022) studied the retraining of AlexNet, ResNet18, and GoogleNet architectures using the fine-tuning method with BCCD datasets from Kaggle, employing SoftMax and SVM for classification. This approach demonstrated that the hybrid architecture combining ResNet18 with SVM achieved an accuracy of 99.83%. Ahmad et al. (2023) proposed extracting optimal deep features from enhanced and segmented WBC images using DenseNet201 and Darknet53 via transfer learning. They employed an entropy-controlled marine predator algorithm (ECMPA) to filter and select dominant features from a serially fused feature vector. The optimized feature vector was then classified using multiple baseline classifiers with different kernel settings. In this line, Wang et al. (2024) applied an unsupervised ML technique combining dimensionality reduction and eigenvector analysis to explore clinical feature relationships. They discovered that baseline serum WBC counts were predictive of overall survival, revealing a significant median survival difference of over six months between the highest and lowest quartiles. Moreover, they observed increased PD-L1 expression in glioblastoma patients with higher WBC counts, identified using an objective PD-L1 quantification algorithm.

In recent years, DL has become increasingly prevalent in medical image classification, particularly for WBC identification. For instance, Thanh et al. (2018) employed CNNs to distinguish between two WBC classes from the ALL-IDB dataset, achieving a 96.6% accuracy. Similarly, Macawile et al. (2018) operated CNNs to classify five WBC classes within the same dataset, reaching an accuracy of 96.63%. Liang et al. (2018) developed a hybrid model combining CNN and recurrent neural network (RNN) architectures to categorize four WBC types in the BCCD dataset, resulting in a 90.79% accuracy. Yu et al. (2017) reported an 88.5% accuracy in classifying seven WBC types using CNNs on a proprietary dataset. Remarkably, Jiang et al. (2018) achieved an 83% accuracy in identifying 40 WBC types from a substantial dataset of 92,800 images. Further advancements include the dual attention CellNet (DACellNet) model by Ye et al. (2019), which accomplished fine-grained recognition of 40 WBC classes with minimal human intervention, attaining accuracy of 84.21% on a leukocyte dataset, 98.78% on ALL-IDB, and 94.7% on the Cellavision database. Patil et al. (2021) proposed a model integrating CNN and RNN with canonical correlation analysis, demonstrating a 95.89% accuracy in classifying four WBC types using public data from the Shenggan/BCCD dataset and Kaggle. Khouani et al. (2020) elaborated a DL approach to automatically recognize WBCs in images, improving hematologists' efficiency with advanced preprocessing and enhanced segmentation techniques. Sharma et al. (2022) implemented a DL model using DenseNet121 to classify various types of WBC. The model was optimized with normalization and data augmentation preprocessing methods. It succeeded an accuracy of 98.84%, a precision of 99.33%, a sensitivity of 98.85%, and a specificity of 99.61%. Girdhar et al. (2022) projected an algorithm to detect WBCs from microscope images, utilizing the simple relationship of colors R and B, along with morphological operations. They then employed a granularity feature (pairwise rotation invariant co-occurrence local binary pattern, PRICoLBP) and SVM to initially classify eosinophils and basophils from other WBC types. Dong et al. (2023) innovatively proposed a WBC classification algorithm that combines DL features with artificial features. This algorithm not only utilizes artificial features but also leverages the self-learning capabilities of Inception V3 to fully exploit the image's feature information. Amine Tahiri et al. (2023) investigated a classification method divided into two main phases: preprocessing, which computes image moments using a new quaternion Meixner-Charlier hybrid moment characterized by parameters α, β, and φ, and optimization using the grey wolf algorithm to enhance classification accuracy. Yentrapragada (2023) designed a combined CNN structure merging AlexNet, GoogLeNet, and ResNet-50 to extract 3000 crucial features. In the same way, Li and Liu (2023) proposed a color-invariant trainable convolutional layer for edge-feature extraction, compatible with various backbone architectures. Trained on the Raabin-WBC dataset, the model achieved 97.8% accuracy. In the same line, Ratheesh and Breethi (2024) employed threshold-based segmentation combined with SqueezeNet-based feature extraction, achieving competitive classification performance with substantially fewer parameters than conventional CNN architectures across four datasets; however, on the Raabin-WBC dataset, the reported accuracy was 94.28%. Similarly, to classify leukocytes from blood smear images, Bhatia et al. (2024) employed various pre-trained DL models, such as DenseNet121, Xception, MobileNetV2, ResNet50, and VGG16. Recently, Li et al. (2025) introduced a modified MobileNetV2 architecture, referred to as MobileNetV2-BAM, and evaluated its performance across multiple datasets; however, on the Raabin-WBC dataset, the reported classification accuracy achieving approximately 97.76%.

To enhance the accuracy of WBC classification, researchers have developed hybrid methodologies that integrate various feature extraction techniques with different classifiers. For instance, Zhao et al. (2017) combined granular features and CNNs with SVM and RF classifiers to categorize five WBC classes across datasets from Cellavision, ALL-IDB, and Jiashan, achieving a 92.8% accuracy. Similarly, Hegde et al. (2018) proposed a hybrid classifier merging SVM and neural networks (NN) for five-category WBC classification, resulting in an average accuracy of 96%. In another methodology, Sengur et al. (2019) applied shape and deep features to represent WBCs and devoted a long short-term memory (LSTM) network to classify four WBC types, attaining an 85.7% accuracy. Furthermore, Sahlol et al. (2020) applied an efficient WBC leukemia classification method that incorporated improved swarm optimization of deep features on two datasets, achieving accuracies of 96.11% and 83.3%, respectively. Özyurt (2020) employed AlexNet, VGG-16, GoogleNet, and ResNet as feature extractors, combining outputs from their last layers. Efficient features selected via minimum redundancy maximum relevance were classified using an extreme learning machine (ELM) classifier, diverging from typical CNN approaches. Baghel et al. (2022) described an automatic classification method using machine learning to classify blood cells from medical images of blood samples. Similarly, Ha et al. (2022) proposed a novel semi-supervised WBC classification method called fine-grained interactive attention learning (FIAL). It features a semi-supervised teacher-student (SSTS) module and a fine-grained interactive attention (FGIA) mechanism, using limited labeled WBC images to generate predicted probability vectors for numerous unlabeled WBC samples. Zhu et al. (2023) introduced DLBCNet for blood cell multi-classification, employing a synthetic image-generating model (BCGAN) and a pre-trained ResNet50 as the feature extractor. The suggested ETRN enhances multi-classification, achieving average accuracy, sensitivity, precision, specificity, and f1-score of 95.05%, 93.25%, 97.75%, 93.72%, and 95.38%, respectively. Aksoy (2024) proposed a hybrid model that achieved 95.60% accuracy and an F1-score of 95.0%, indicating a potentially less optimized or less diverse approach compared to other methods.

Model-agnostic interpretation techniques, often referred to as explainable artificial intelligence (XAI) models, gained popularity in 2017 for their ability to provide transparency into AI models' results by revealing their underlying mechanisms and logic. Various XAI models have been obtained, including instance-based methods like LIME, feature-based methods like SHAP, and gradient-based techniques such as DeepLIFT and Attribution Maps. XAI has revolutionized the healthcare industry by enabling medical professionals and researchers to gain a deeper understanding of AI systems and their decision-making processes. This transparency has made AI systems more accountable and reliable, fostering greater trust among healthcare providers in using AI-based recommendations. Most XAI algorithms are post-hoc and model-agnostic, meaning they can be applied to any black-box model (Ribeiro et al. 2016; Korica et al., 2021; Islam et al., 2024). In 2025, Dagnaw et al. (2025) presented a comprehensive and structured review of XAI methods for biomedical imaging, where they systematically categorize techniques and analyze their principles, strengths, and limitations.

### 2.2 Research gap and contributions

The literature review indicates that, despite the availability of various diagnostic and classification approaches for WBCs, several challenges remain, including complex model configurations, increased computational complexity in some studies, and limitations in classification accuracy. While many existing WBC diagnosis systems report reasonable performance in distinguishing the five major WBC classes, relatively few studies focus on simultaneously improving classification accuracy and computational efficiency for multi-class WBC classification. Several ensemble-based deep learning approaches have been proposed, such as ensemble-fused CNN architectures (Banik et al., 2019), hybrid ensemble frameworks integrating CNN and RNN models via canonical correlation analysis (Patil et al., 2021), ensemble fine-tuning of multiple pre-trained CNN architectures including AlexNet, ResNet18, and GoogleNet (Cengil et al., 2022), and bagging-based ensembles combining VGG16, ResNet50, and Inception-V3 with parallel training (Dong et al., 2024). These studies demonstrate the effectiveness of ensemble learning for WBC classification; however, the associated increases in computational complexity and inference time highlight the need for more balanced ensemble designs that achieve high accuracy while maintaining practical efficiency. Although some hybrid-based models have shown superior accuracy on comprehensive five-class datasets, their high computational complexity poses challenges for real-time applications. Our focus is to achieve high classification accuracy, particularly for critical WBC classification tasks. In predicting WBC classes, XAI is often viewed as a black-box approach, utilizing a linear model to determine the relative importance of different sample features. To address these challenges, we aim to develop effective learning strategies that optimize computational efficiency through customized CNN layers.

In our research work, we proposed a robust WBC classification technique named the deep three CNN ensemble network for WBC classification (DCENWCNet). To achieve this, we introduce an combination model formed upon a set of custom-built deep CNNs, each designed with specific learning constraints. Such constraints facilitate feature extraction at multiple levels by incorporating different dropout configurations. Specifically, the three deep CNNs are designed with progressively decreasing dropout depths (three, two, and one dropout layers, respectively), each integrated with corresponding max-pooling layers to introduce controlled regularization, promote architectural diversity, and enhance complementary feature learning across the ensemble. Initially, we performed a comparative analysis of various pre-trained networks on the Raabin-WBC dataset to assess their effectiveness in blood cell classification. Based on these findings, we developed three hybrid models. The first model integrates feature representations from three customized CNNs, which are subsequently fed into a classifier for final classification. To enhance performance, we designed the DCENWCNet model, leveraging a hybrid approach that combines multiple networks with learning constraints. These constraints regulate feature extraction through varying dropout and max-pooling configurations, enabling effective learning at different levels. While ensemble models are generally known for their superior accuracy, their computational demands pose challenges for real-time applications. However, our main objective is to maximize classification accuracy, particularly in sensitive tasks such as WBC detection, while also exploring optimized learning techniques such as adjusting dropout rates and max-pooling strategies to reduce computational overhead. Furthermore, we applied pixel standardization, strategic data augmentation, and a well-balanced annotated dataset to improve the model's reliability and generalization capabilities. To assess the robustness and generalization capabilities of our methodology, we conducted comparative analyses against established ML models, including VGG19, VGG16, Inception-V3, ResNet-50, and SVM. These evaluations were performed on both our proprietary dataset and the publicly available Raabin-WBC dataset. We evaluated performance using metrics such as precision, recall, specificity, F1-score, and the area under the receiver operating characteristic curve (AUC-ROC). Ultimately, taking these design, testing, and validation aspects into account, we propose a structured ensemble framework that leverages heterogeneous dropout-based regularization strategies across parallel CNN branches to improve robustness, generalization, and interpretability in WBC classification, with the detailed methodology outlined in the following section. The contribution of this work lies in the controlled introduction of architectural diversity through regularization-driven CNN ensembles, rather than in proposing a new ensemble learning paradigm.

## 3. Preliminary materials and proposed methodology

In this study, our aim is to develop an efficient method which classifies WBC images into five classes. Thus, for classification purposes, we provide a brief overview of require deep neural networks in the following subsection.

### 3.1 Convolution neural network (CNN)

Currently, in the area of computer vision, various CNN pre-trained models e.g., GoogLeNet (Szegedy et al. 2015), ResNet (He et al. 2016), AlexNet (Krizhevsky et al. 2017), and VGGNet (Simonyan & Zisserman, 2014), among others, have been introduced. Such kind models like GoogLeNet and ResNet, are also accessible as pre-trained on around 1.28 million natural images from the ImageNet database. This allows us to utilize the pre-established weights and biases of these models. By adjusting the layers of these models through backpropagation with our specific data, they can be adapted for our unique classification tasks. Conversely, pre-trained models e.g., ResNet and VGGNet are designed with initial weights and biases that remain unaffected by pictorial information not related to our predicted images. Here, we provide a concise summary of these widely utilized CNN architectures.

### 3.2 Proposed methodology

The proposed ensemble consists of three base networks with progressively decreasing dropout depths. This design introduces controlled architectural diversity while maintaining computational efficiency. The use of varying dropout depths enables the ensemble to capture complementary feature representations at different regularization levels, which is particularly important for handling intra-class variability in WBC images. The models were concurrently trained using the augmented image sets via a sequence of convolutional layer, max pooling, and dropout layers.

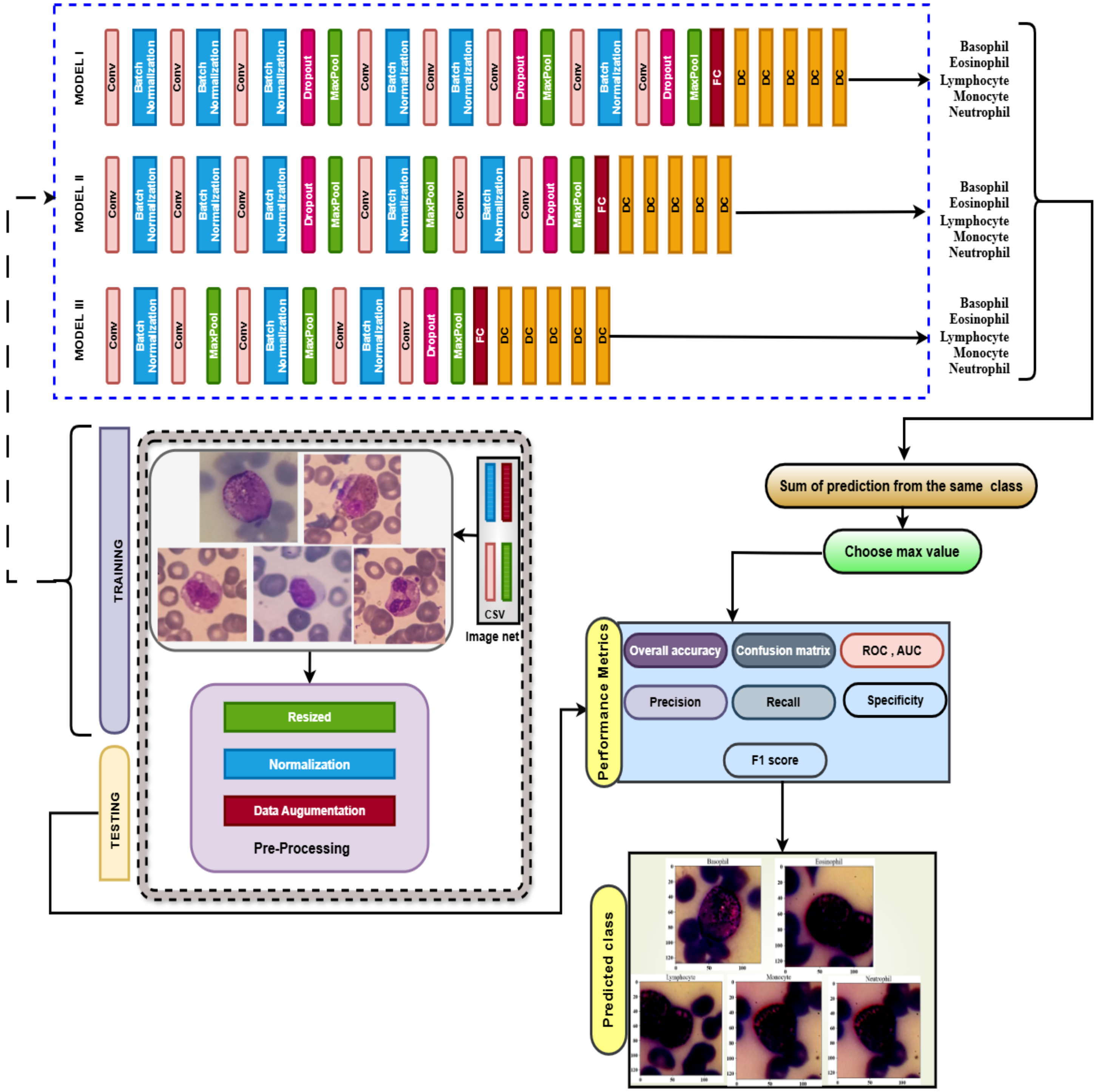

Fig. 2: The workflow of the WBC classification proposed model DCENWCNet

Unlike conventional ensembles that aggregate independently trained models without structural intent, the proposed DCENWCNet enforces diversity through progressive regularization depth. Model-I emphasizes high-capacity learning, Model-II enforces moderate regularization, and Model-III promotes generalization through aggressive regularization. This structured diversity ensures complementary error patterns, which the probabilistic aggregation exploits. Fig. 2 displays the whole design, with detailed technical information for each level provided thereafter.

### 3.3 Proposed design

The DCENWCNet architecture employs a compact ensemble of three DCNNs to aggregate the prediction outputs of individual models and produce the final class decision. The choice of three ensemble models was guided by a deliberate balance between representation diversity, regularization strength, and computational efficiency, rather than exhaustive ablation over all possible ensemble sizes. Preliminary exploratory experiments indicated that growing the number of parallel models beyond three resulted in marginal performance gains while substantially increasing training time and computational complexity. All three networks are trained simultaneously on the augmented image dataset and share a common backbone structure, differing only in their dropout configurations. Specifically, the networks incorporate progressively decreasing numbers of dropout layers, which enables controlled regularization diversity across ensemble models. This design encourages complementary feature learning by allowing different levels of model expressiveness without introducing heterogeneous architectures. The integration strategy of the proposed framework is summarized in Table 1. Each individual model produces a one-dimensional probability vector corresponding to the five WBC classes, where each element represents the confidence score for a specific class. The outputs of the three models are then stacked to form a two-dimensional array, and the class-wise confidence values are aggregated across models through summation. The final classification result is determined by choosing the class corresponding to the highest aggregated confidence score, where all utilized symbols and parameters are explicitly defined in Table 2.

**Table 1:** Prediction procedure of combining models

| Step | Procedure |
|---|---|
| 1. | Input: Provide the trained CNN models [MODEL I,MODEL II,MODEL III]and the test dataset. |
| 2. | Output Definition: Define the output as the ensemble prediction. |
| 3. | Model Set Formation: Initialize the ensemble model list as [MODEL I,MODEL II,MODEL III]. |
| 4. | Prediction Loop: For each model in the ensemble, perform prediction on the test dataset. |
| 5. | Probability Estimation: Generate class-wise probability predictions using the Softmax output of each CNN model. |
| 6. | Loop Termination: Complete prediction generation for all models. |
| 7. | Prediction Array Construction: Store all model-wise probability predictions in a single array. |
| 8. | Probability Summation: Compute the sum of predicted probabilities across all models for each class. |
| 9. | Final Decision Rule: Select the class with the maximum aggregated (normalized) probability using the argmax operation. |
| 10. | Final Prediction: Assign the selected class as the ensemble prediction. |

**Table 2:** Definition of symbols and parameters

| Symbol | Description | Dimension / Type |
|---|---|---|
| IM | Input WBC image | $\xi \times \psi \times \omega$ |
| $\xi$, $\psi$ | Height and width of the input image | Pixels |
| $\omega$ | Number of image channels (RGB) | $\omega = 3$ |
| *KR* | Convolution kernel (filter) | $l_1 \times l_2 \times \omega \times D$ |
| $l_1$, $l_2$ | Kernel height and width | Pixels |
| *D* | Number of convolution filters | Integer |
| i, j | Spatial indices of output feature map | Integer |
| s, t | Kernel spatial indices | Integer |
| r (L) | Channel (Layer) index | Integer (+ve integer) |
| $x_{i,j}^{L}$ | Feature map value at layer L | Scalar |
| $x^{L-1}$ | Feature map from previous layer | Tensor |
| $W^{L}$ | Weight matrix of layer L | Learnable parameters |
| b ($\alpha$) | Bias term (vector) | Scalar (vector) |
| $\Omega(\cdot)$ | Activation function (ReLU) | Nonlinear |
| *ER* | Error term | Scalar |
| $\partial/\partial(\cdot)$ | Partial derivative operator | — |
| *N* | Pooling window size | N × N |
| $\max(\cdot)$ | Max pooling operator | — |
| $x_s$, $y_s$ | Pooled output values | Scalar |
| $\xi_{x,y}^{L}$ | Feature value after max-pooling | Real-valued |
| $\psi_r(\cdot)$ | ReLU activation function | max(0,r) |
| $\psi_\omega(\cdot)$ | Softmax activation function | Probability mapping |
| $\omega_i$ | Logit value for class $i$ | Real-valued |
| $P_i$ | Predicted probability of class $i$ | [0, 1] |
| *n* | Number of classes | *n=5* |
| *m* | Number of CNN models in ensemble | *m=3* |
| $P_{i,j}$ | Probability of class $i$ from model $j$ | [0, 1] |
| $\widehat{P_i}$ | Normalized aggregated class probability | [0, 1] |
| $C_i$ | Class label | Categorical |
| TP (TN) | True Positives (True Negatives) | Integer |
| FP (FN) | False Positives (False Negatives) | Integer |
| TPR | True Positive Rate | TP / (TP + FN) |
| FPR | False Positive Rate | FP / (FP + TN) |
| $\mu$ | Mean pixel intensity | Scalar |
| $\sigma$ | Standard deviation | Scalar |
| $x'$ | Normalized pixel value | Scalar |

*(i) Convolutional layer*

Our deep CNN model is structured with a series of convolutional filters designed to capture and extract essential features from the images. Given an input image $IM \in \mathbb{R}^{\xi\times\psi\times\omega}$and a convolution kernel $KR \in \mathbb{R}^{l_1\times l_2\times\omega\times D}$with spatial dimensions $l_1 \times l_2$, the convolution operation is formulated in Eq. (1). Consequently, for RGB images containing three color channels ($\omega = 3$), the output of the convolutional layer network is described in Eq. (2), where $\alpha$ represents the bias vector i.e., i.e., $\alpha \in \mathbb{R}^D$ corresponding to the filter bank of size $D$.

$$\begin{aligned}(IM * KR)_{i,j} &= \sum_{s=0}^{l_1-1}\sum_{t=0}^{l_2-1} IM(i-s, j-t)KR(s,t) \\ &= \sum_{s=0}^{l_1-1}\sum_{t=0}^{l_2-1} IM(i+s, j+t)KR(-s,-t)\end{aligned} \tag{1}$$

$$(IM * KR)_{i,j,r} = \sum_{s=0}^{l_1-1}\sum_{t=0}^{l_2-1}\sum_{r=1}^{\omega} KR_{s,t,\omega}.IM_{i+s,j+t,r} + b \tag{2}$$

As the network progresses through successive convolutional layers networks, the mapping between two adjacent layers is formulated in Eq. (3). In this formulation, $\Omega(\cdot)$ denotes the activation function, $W^L$ represents the weight parameters of the $L$-th layer, and $x^L$ denotes the corresponding output feature map at that layer.

$$x_{i,j}^{L} = \sum_{s}\sum_{t} W_{s,t}^{L}\Omega(x_{i,j})_{i+s,\mathrm{j+t}}^{L-1} + b^{L} \tag{3}$$

The activation function is first used to the output of the earlier layer $x^{L-1}$, after which the result is weighted and combined with the corresponding bias terms. The network parameters are then updated through backpropagation across the convolutional layers, as described in Eq. (4), where the error term $ER$is evaluated at the output layer. The resulting output $x_{i,j}^L$ is subsequently defined in Eq. (5).

$$\frac{\partial ER}{\partial W_{s',t'}^{l}} = \sum_{i=0}^{\xi-l_1}\sum_{j=0}^{\psi-l_2} \frac{\partial ER}{\partial x_{i,j}^{L}} \frac{\partial x_{i,j}^{L}}{\partial W_{s',t'}^{l}} \tag{4}$$

$$x_{i,j}^{L} = \sum_{s}\sum_{t} W_{s,t}^{L}\Omega(x_{i,j})_{i+s,\mathrm{j+t}}^{L-1} + b^{L} \tag{5}$$

*(ii) Max pooling layers*

It is usually used for feature dimensionality reduction to achieve the purpose of compressing data and reducing the number of parameters in the training process, thereby reducing overfitting and improving the fault tolerance of the model. There are two main types of pooling layers: maximum pooling and average pooling. This article uses maximum (Max) pooling. The proposed model incorporates pooling layers with a 2 × 2 pooling size. The max-pooling operation is adopted in place of average pooling, as average pooling tends to smooth feature maps and may suppress important high-intensity or edge-related features. In contrast, max pooling preserves salient local responses by retaining the maximum activation within each pooling region. The max-pooling operation applied to an $N \times N$ region of the feature map is stated in Eq. (6), where $s_x$ and $s_y$ denote the values at spastial locations $(x, y)$ after pooling, and $\overline{s}_x$ and $\overline{s}_y$ represent the corresponding values prior to the pooling operation.

$$\xi_j^L\left(s_x, s_y\right) = \underset{\overline{s}_x \in N, \overline{s}_y \in N}{Max}\left(\overline{s}_x, \overline{s}_y\right) \tag{6}$$

*(iii) Dropout regularization*

The dropout layer achieves the effect of using random deactivation of hidden units to prevent overfitting. The introduction of this randomness forces the network to become redundant, so that the network does not match the training samples well, thereby increasing the generalization ability of the network of the proposed model. These three models within the proposed ensemble manner are primarily differentiated by their layer configurations and the number of dropout layers utilized. Model-I incorporates three dropout layers, while Model-II reduces this to two dropout layers. In contrast, Model-III includes one dropout layer. This deliberate variation in dropout layer design aims to strike a balance between bias and variance trade-offs, facilitating more effective learning. Even if one model overfits to a specific set of images, the presence of the other two models helps mitigate this issue, ensuring overall robustness and generalization.

*(iv) Batch normalization*

We have incorporated batch normalization layers network between the convolutional layers to standardize the input for the subsequent layer. This process involves scaling and shifting the data received from the previous layer network. Notably, the model learns the scaling and offset coefficients through backpropagation.

*(v) Dense layers*

This layer comprises a vast number of neurons that establish connections between different layers. Fully connected layers, combined with activation functions, play a crucial role in the network. Specifically, the flattening manner restructures the multidimensional output from the convolutional layer into a one-dimensional vector. Each of the three models contains four fully connected hidden layers, followed by a final layer with five neurons corresponding to the five image classes. The outputs from this final layer are subsequently integrated into the ensemble process. When the dense layer employs the Softmax activation function, it functions as a classification layer.

*(vi) Hyperparameters*

The hyperparameters of the proposed model, which represent the estimated computational complexity of the neural network (NN), are outlined in Table 3. In this architecture, the hidden layers utilize the Rectified Linear Unit (ReLU) activation function, while the output layer employs the Softmax activation function, as specified in Eqs. (7) and (8), respectively.

$$\Psi(r) = r^+ = \max(0, r) \tag{7}$$

$$\Psi(\omega) = \frac{e^{w_i}}{\sum_i e^{w_i}}, \quad \forall \omega_i \in W \tag{8}$$

Additionally, we have utilized the well-known optimization techniques such as Adam and RMSprop.

### *3.4 Theoretical framework of the hybrid models*

For hybrid CNN models, we focus on the outputs of their classification layers which take the input from the fully connected layers and calculate confidence values for each classification (in our proposed model, we have n = 5 classes). For the adequate formalization, we consider a CNN to be a function $\psi: x \rightarrow \mathbb{R}^n$ that assigns n confidence values $P_i \in [0,1]$ for $i = 0,1,...,n,$ & $\sum_{i=0}^{n} P_i = 1$. In our proposed model, $P_0, P_1, P_2, P_3, P_4$ denoted the confidence of the given CNN models that $x$ should be classified as lymphocytes $(C_0)$, monocytes $(C_1)$, neutrophils $(C_2)$, eosinophils $(C_3)$, and basophils $(C_4)$, respectively. As a simple conclusion, the CNN assigns x to the class having the maximal probability:

$$x \rightarrow C_i \text{, if } P_i = \max(\psi(x)) \tag{9}$$

In our proposed model, the CNNs classify 5 confidence values to each test image to give probability that the image was labeled by monocytes, eosinophils, basophils, lymphocytes, and neutrophils, respectively. Therefore, need to drive $P_i^{/}$ probabilities for each image from the confidence score of the individual trained model DCENWCNet in out hybrid models, where $P_i^{/} \in [0,1]$, $i = 0,1,...,n,$ & $\sum_{i=0}^{5} P_i = 1.$ The possible confidence values are aggregated using maximum of probabilities (MOP) prediction approaches are discussed next.

*(i) Max of probability (MOP)*

The proposed ensemble framework aggregates class-wise confidence scores produced by individual CNN models using a normalization-aware probability summation strategy. First, each CNN model generates Softmax-based posterior probabilities for all classes. For a given input image, the predicted probabilities corresponding to the same class are summed across all ensemble models. The aggregated class-wise scores are then normalized to ensure probabilistic consistency. Finally, the class with the maximum normalized aggregated probability is selected as the final prediction. Unlike simple voting or unnormalized probability averaging, the MoP strategy preserves class-wise confidence magnitudes while ensuring that the aggregated probabilities form a valid distribution. This enables robust and consistent decision-making across models with potentially different uncertainty characteristics. In Fig. 3, each summation of the prediction scores is shown in 0.8, 1.5, 1.1, 1.3, 1.6 for class lymphocytes, monocytes, neutrophils, eosinophils, and basophils, respectively. The normalization step make certain that the aggregated probabilities sum to one, and the image class associated with the maximum normalized value is selected as the final prediction. Based on existing literature (Jacobs, 1995); (Kittler et al., 1998), the MOP aggregation process for the proposed architecture is formulated in Eq. (10), where i and j are the index for the class and DCENWCNet models. The MOP aggregation procedure for the proposed network is expressed in Eq. (1), where $i$ denotes the class index and $j$ denotes the DCNN model index. Accordingly, $P_{i,j}$ represents the prediction score for the $i$ th class (among n total classes) generated by the $j$th model (among m total models). A normalization factor, $\sum_{i=1}^{n}\sum_{j=1}^{m} P_{i,j}$, is applied to scale the values after summing the prediction scores for each class, $\sum_{j=1}^{m} P_{i,j}$.

$$P_i^{/} = \arg\max_i \left( \frac{\sum_{j=1}^{m} P_{i,j}}{\sum_{i=1}^{n}\sum_{j=1}^{m} P_{i,j}}, \forall i \right) \tag{10}$$

The term, normalization $\sum_{i=1}^{n}\sum_{j=1}^{m} P_{i,j}$ is applied to have $P_i^{/} \in [0,1]$, $i = 0,1,\dots,n$, & $\sum_{i=0}^{5} P_i = 1$.. In the proposed model, we initially compute the sum of confidence values from each trained model $CNN_j$ (j = 1, ..., m) for every class independently, and then determine the image's final label by choosing the maximum normalized sum.

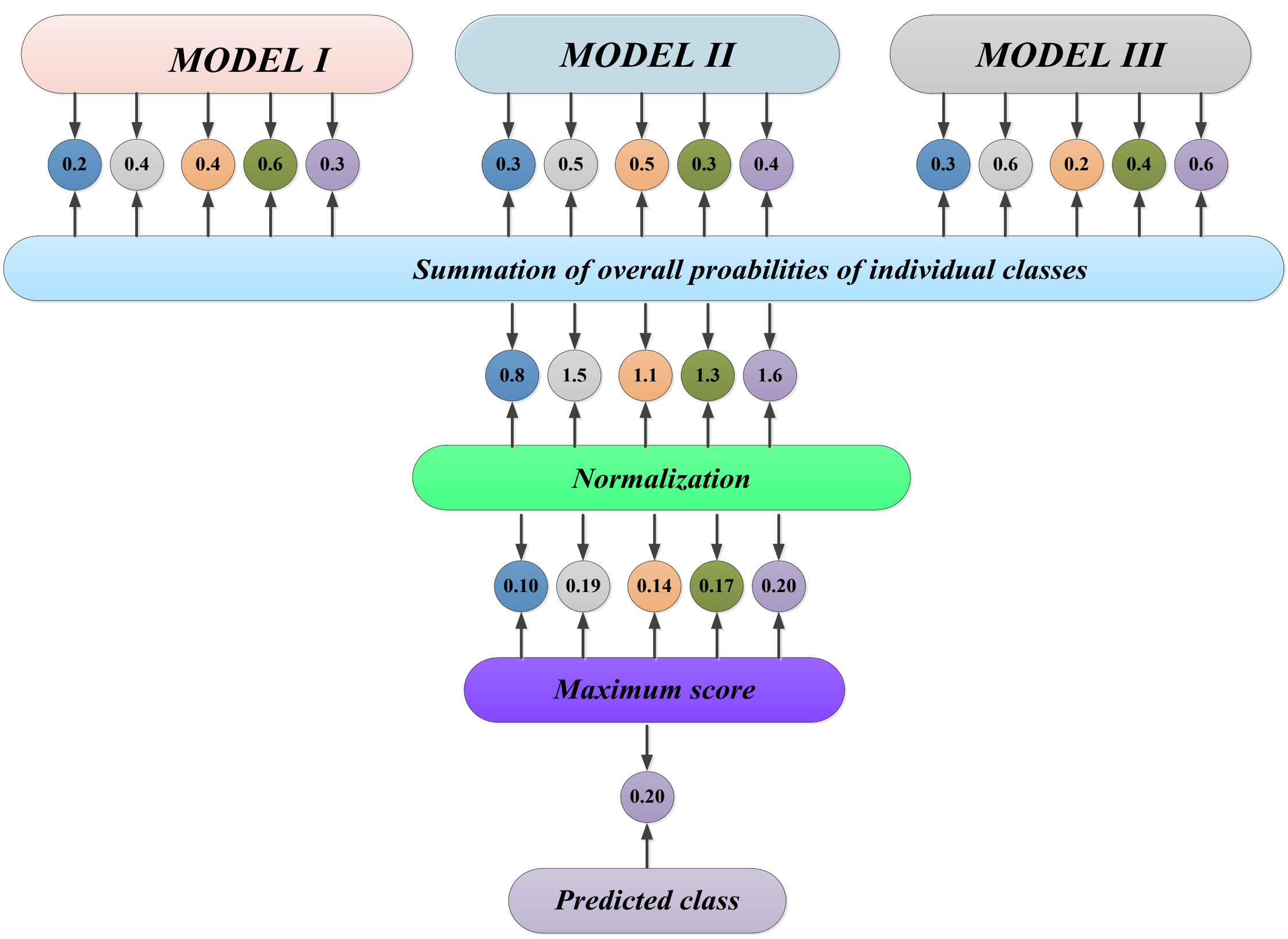

Fig. 3: Proposed hybrid model utilizing MOP for class prediction example

## 4. Data experimental and environment setup

We now describe the experimental setup used to train, test, and validate this research model DCENWCNet framework for white blood cell (WBC) classification. Each method was executed over 30 independent runs during the training phase to ensure reliability and reduce the influence of random initialization. The key technical choices, together with the mathematical formulation presented earlier, are organized into an end-to-end workflow (Fig. 5). This workflow is divided into three major stages: (i) dataset preprocessing, (ii) model training and validation, and (iii) final WBC class prediction. The following subsections briefly summarize these stages and highlight the experimental configurations required to rigorously evaluate classification performance.

### 4.1 Data description

The experiments in this study use the Raabin-WBC dataset (Rubin et al., 2023). It is a large, curated collection of 14,514 labeled images gathered from multiple sources. The dataset covers five major leukocyte categories: (i) basophils, (ii) eosinophils, (iii) lymphocytes, (iv) monocytes, and (v) neutrophils. Table 3 summarizes representative medical conditions linked to abnormal variations in WBC counts. For example, allergic responses are commonly associated with elevated basophil levels, whereas hematological cancers may increase immature cell precursors and produce noticeable changes in cell morphology, including size and shape. Therefore, reliable identification of WBC type together with accurate counting plays an essential role in supporting clinical diagnosis across a broad range of disorders.

### 4.2 Data pre-processing

For WBC recognition, this study uses the widely adopted Raabin-WBC dataset. It contains 14,514 images for model development and reflects considerable variability in subject characteristics (e.g., age, sex, and ethnic background), which helps the evaluation better represent real-world conditions. The images are grouped into five leukocyte categories, and the number of samples in each class is reported in Table 3. Before training, the dataset is prepared through three main preprocessing steps: (i) class-balancing via controlled sampling, (ii) pixel-value standardization, and (iii) data augmentation to improve generalization. The details of these procedures are provided in the following subsections, together with the relevant implementation settings.

**Table 3:** Sample images from the Raabin-WBC dataset accompanied with comprehensive metadata

| Labels | Types | Sample Images | # of Images | Description |
|---|---|---|---|---|
| 0 | Basophil | | 301 | ***Decrease:*** Hyperthyroidism and acute infections<br>***Increase:*** Leukemias |
| 1 | Eosinophil | | 1066 | ***Decrease:*** Cushing syndrome, shock or trauma driven stress.<br>***Increase:*** Allergic reaction, parasitic infection, malignancy |
| 2 | Lymphocyte | | 3461 | ***Decrease:*** AIDS, influenza, sepsis, aplastic anemia.<br>***Increase:*** Acute and chronic leukemia, hypersensitivity reaction, viral infection |
| 3 | Monocyte | | 795 | ***Decrease:*** Aplastic anemia, hairy cell leukemia, acute infections.<br>***Increase:*** Autoimmune disease, fungal and protozoan infection. |

| 4 | Neutrophil | 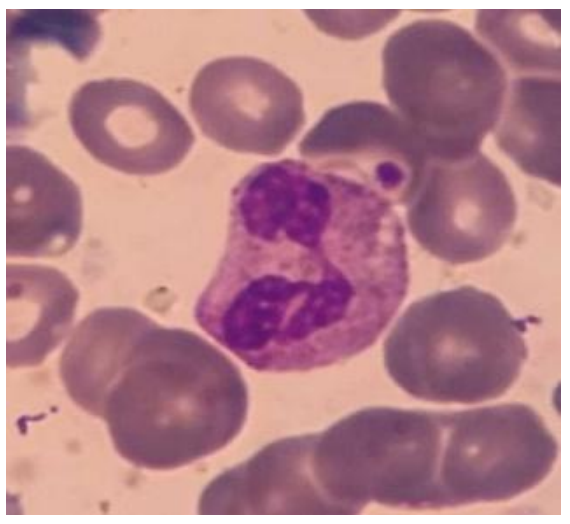 | 8891 | ***Decrease:*** Chediak-Higashi syndrome, Kostmman syndrome, Autoimmune neutropenia. ***Increase:*** Chronic inflammation, Infection |
|---|---|---|---|---|

*(i) Dataset balancing*

The Raabin-WBC dataset is organized in a CSV file (Rubin et al., 2023) comprising 17,965 rows, where each row corresponds to a single microscopic image and contains 2,353 columns, with the final column indicating the associated class label. The dataset includes a total of 14,514 white blood cell images distributed across five classes. For model training, 80% of the dataset is utilized, consisting of 212 Basophils, 744 Eosinophils, 2,427 Lymphocytes, 561 Monocytes, and 6,231 Neutrophils, reflecting a pronounced class imbalance between minority and majority classes. The remaining 20% of the data is reserved for testing and includes 89 Basophils, 322 Eosinophils, 1,034 Lymphocytes, 234 Monocytes, and 2,660 Neutrophils. Detailed class distributions and representative samples for both training and testing sets are presented in Table 3. All images are represented as 64 × 64 × 3 RGB pixel arrays and reshaped accordingly to form the final input images. The compact spatial resolution is selected to reduce computational complexity while preserving discriminative features, leveraging the inherent scale and translation invariance of CNN backbone architectures. This representation supports efficient training and contributes to stable learning across classes prior to the application of data augmentation strategies described in the subsequent subsection.

*(ii) Data pixel standardization*

Prior to feeding the training images into the model, data normalization is performed using the widely adopted Gaussian pixel standardization technique. As defined in Eq. (13), this process normalizes each pixel by subtracting the mean pixel intensity and dividing by the standard deviation computed over the training dataset. This normalization ensures consistent feature scaling across images, thereby improving learning stability and convergence during training. Gaussian pixel standardization also reduces the influence of noise, outliers, and extreme pixel values through centralization, which is particularly beneficial in image classification tasks where distance-based similarity implicitly affects feature learning. Consequently, this normalization strategy facilitates more stable optimization by mitigating issues such as exploding gradients that may arise from unscaled pixel intensities. Owing to these advantages, pixel standardization is adopted in this study. Representative examples of the standardized Raabin-WBC images are provided in Table 4.

**Table 4:** Sample images of Raabin-WBC dataset: original and with augmentation and pixel-standardization

| Types | Original | After augmentation and pixelization | | | |
|---|---|---|---|---|---|
| | | Aug: 1 | Aug: 2 | Aug: 3 | Aug: 4 |
| *Baspphil* | | | | | |
| *Eosinophil* | | | | | |
| *Lymphocyte* | | | | | |
| *Monocyte* | | | | | |
| *Neutrophil* | | | | | |

*(iii) Data augmentation*

Before being utilized for training, the Raabin- WBC dataset undergoes data augmentation to ensure a more balanced distribution among its categories. This process helps mitigate the risk of bias towards classes with a higher count of training images, which would otherwise likely achieve greater accuracy compared to classes with fewer images. For the augmentation of images, the dataset leveraged an image data generator within Keras preprocessing, a component of TensorFlow. The specifics of the augmentation process are detailed in Table 5. As a result of augmentation, the image count in the dataset expanded from 14,514 to 50,024, significantly enlarging the training dataset. Given the pronounced class imbalance in the original Raabin-WBC dataset, balanced sampling and class-aware data augmentation were adopted to increase the effective representation of minority classes during training. This strategy reduces bias toward dominant classes and enables stable learning without requiring class-weighted loss functions. In addition, ensemble regularization through heterogeneous dropout configurations further improves robustness to class imbalance. This expansion helps reduce the risk of overfitting, as depicted in Fig. 5.

**Table 5:** Image augmentation

| Process Name | Value |
|---|---|
| Rescale | 1/255 |
| featurewise_center | False |
| featurewise_std_normalization | False |
| Rotation Range | 10 |
| Width Shift Range | 0.2 |
| Height Shift Range | 0.2 |
| Shear Range | 0.2 |
| zoom_range | 0.1 |
| Horizontal Flip | True |
| Vertical Flip | True |
| zca_whiteninge | False |
| Fill Mode | nearest |

### 4.3 Software for experiment

For DCENWCNet investigation, we conducted all comparisons on the same CPU machine and same dataset. The runtime of all methods was recorded and compared in the experimental findings. To assess classification performance, we employed the Python-Sklearn library's classification report utility. All experiments in this proposed work were conducted using Python 3.10.0 within a Jupyter Notebook environment. The hardware employed was an HP laptop equipped with an Intel Core i5 processor operating at 3.20 GHz, 8 GB of RAM, and a 4 GB graphics card. For the implementation of our models, we utilized the TensorFlow and Keras libraries.

### 4.4 Performance metrics

Performance metrics are essential for measuring the performance of machine learning and deep learning models. ss

- TP: normal is appropriately classified as normal.
- TN: WBC is appropriately classified as WBC.
- FP: WBC is inappropriately classified as normal.
- FN: normal is inappropriately classified as WBC.

In this proposed model DCENWCNet, we concentrate on various key performance metrics to evaluate the efficiency of the proposed research model. Given the inherent class imbalance of the Raabin-WBC dataset, performance evaluation emphasizes class-wise precision, recall, F1-score, and AUC rather than accuracy alone, ensuring reliable assessment of minority-class performance.

I. *Accuracy*: Assesses the model's overall prediction accuracy by determining the ratio between of samples correctly classified out of the total sample count. Reliance on accuracy alone may not provide a comprehensive evaluation, particularly in cases involving imbalanced datasets. However, accuracy is not only always sufficient condition for evaluation, especially when datasets become imbalanced.

$$Accuracy = \frac{(TP + TN)}{(TP + FP + TN + FN)} \tag{11}$$

II. *Precision*: Precision measures is the proportion of true positives out of the total number of predicted positives, summing true positives and false positives. It emphasizes the trustworthiness of positive predictions.

$$\Pr ecision = \frac{TP}{(TP + FP)} \tag{12}$$

III. *Recall*: This quantity is calculating by the ratio between true positives in relation to the total of true positives and false negatives. It concentrates on the thoroughness of positive predictions.

$$Recall = \frac{TP}{(TP + FN)} \tag{13}$$

IV. *F1 Score*: The F1 score is the harmonic mean of precision and recall, ranging between 0 and 1, where 1 indicates the highest possible performance.

$$F1\,score = \frac{TP}{\left[TP + \frac{1}{2}\{FP + FN\}\right]} \tag{14}$$

V. *Specificity*: It is defined as the proportion of true negative in relation to the total of true negative and false positives. It is denoted as:

$$Specificity = \frac{TN}{(TN + FP)} \tag{15}$$

VI. *AUCROC Score*: It is defined by plotting the true positive rate (TPR) against the False Positive Rate (FPR) at different decision thresholds.

$$TPR = \frac{TP}{TP+FN} \tag{16}$$

$$TPR = \frac{FP}{FP+TN} \tag{17}$$

The ROC curve is obtained by plotting TPR vs. FPR for varying classification thresholds.

The area under the ROC curve (AUC) is a scalar summary metric representing the classifier's ability to discriminate between classes:

$$AUC = \int_0^1 TPR(FPR)\, d(FPR) \tag{18}$$

In a multi-class setting, accuracy measures the share of samples that are assigned to the correct category out of all predictions made. In contrast, precision, recall, and the F1-score are commonly summarized across classes using weighted averaging, where each class contributes according to its number of samples. This reporting strategy is particularly useful when the dataset is imbalanced, because it reflects the unequal representation of classes in the overall evaluation.

## 5 Results and discussion

This section reports the experimental results, beginning with a statistical examination of the predicted labels to evaluate overall classification effectiveness. The discussion is organized into the following subsections:

- Subsection 5.1 display the model's learning progress over epochs, highlighting trends in both loss reduction and accuracy improvement.
- Subsection 5.2 provides in-depth review of the model's classification performance metrices.
- Subsection 5.3 represent a confusion matrix to visualize the model's performance across various classes, aiding in the identification of misclassifications.
- Subsection 5.4 illustrates the model's diagnostic ability by plotting the ROC curve and calculating the AUC, offering a measure of the model's predictive accuracy.
- Subsection 5.5 conducts a comparative analysis of the model's performance against benchmarks of other models.
- Subsection 5.6 compares our proposed model with existing studies

### 5.1 Training and validation accuracy and loss analysis

A well-trained deep learning model is typically reflected by close agreement between training and validation trends, which is commonly inspected through accuracy–epoch and loss–epoch curves. When both curves progress in a similar manner, it indicates that the network is capturing informative structures in the data rather than simply memorizing the training set, and it is therefore more likely to generalize to unseen samples. In contrast, a pronounced gap between training and validation performance—such as high training accuracy with noticeably poorer validation accuracy, or steadily decreasing training loss alongside stagnant or increasing validation loss—signals that the model is fitting the training data too closely, which is characteristic of overfitting.

Therefore, in a well-structured model, accuracy should steadily increase across epochs, while the loss should progressively decline. The training and testing phases should exhibit similar patterns. The key results can obtain as following:

- Figs. 4 (i-ii)-5 (i-ii) present the training and testing accuracy, as well as the training and testing loss, over 50 iterations of the proposed DCENWCNet model on the dataset, utilizing both RMSprop and Adam optimizers, respectively.
- Fig. 5 (ii) demonstrates that the training accuracy appears to surpass the validation accuracy at almost every point, which is a common occurrence as models tend to perform better on data they have seen before.
- The validation accuracy increases alongside the training accuracy, which is a good sign because it means the model is not just memorizing the training data but is also improving its ability to generalize to new data as shown in Fig. 5 (ii)
- We have seen a minor case of overfitting in the loss curve of proposed model (see Fig. 5 (i)), especially noticeable in the elevated loss during the latter phases of training. As illustrated in Fig. 5(i) and (ii), the training and test accuracy curves demonstrate that as the number of iterations increases, the model effectively learns, resulting in successful test accuracy outcomes.

- As depicted in Fig. 5(i) and (ii), both the training and test loss curves demonstrate that as the number of iterations increases, the model effectively learns, leading to a decrease in the test error rate.

Comparison of Fig. 4 and Fig. 5 reveals that the accuracy results of our proposed model using the Adam optimizer are much better compared to the RMSprop optimizer. Therefore, for further performance metrics and classification accuracy comparisons with existing articles, we will only consider the Adam optimizer.

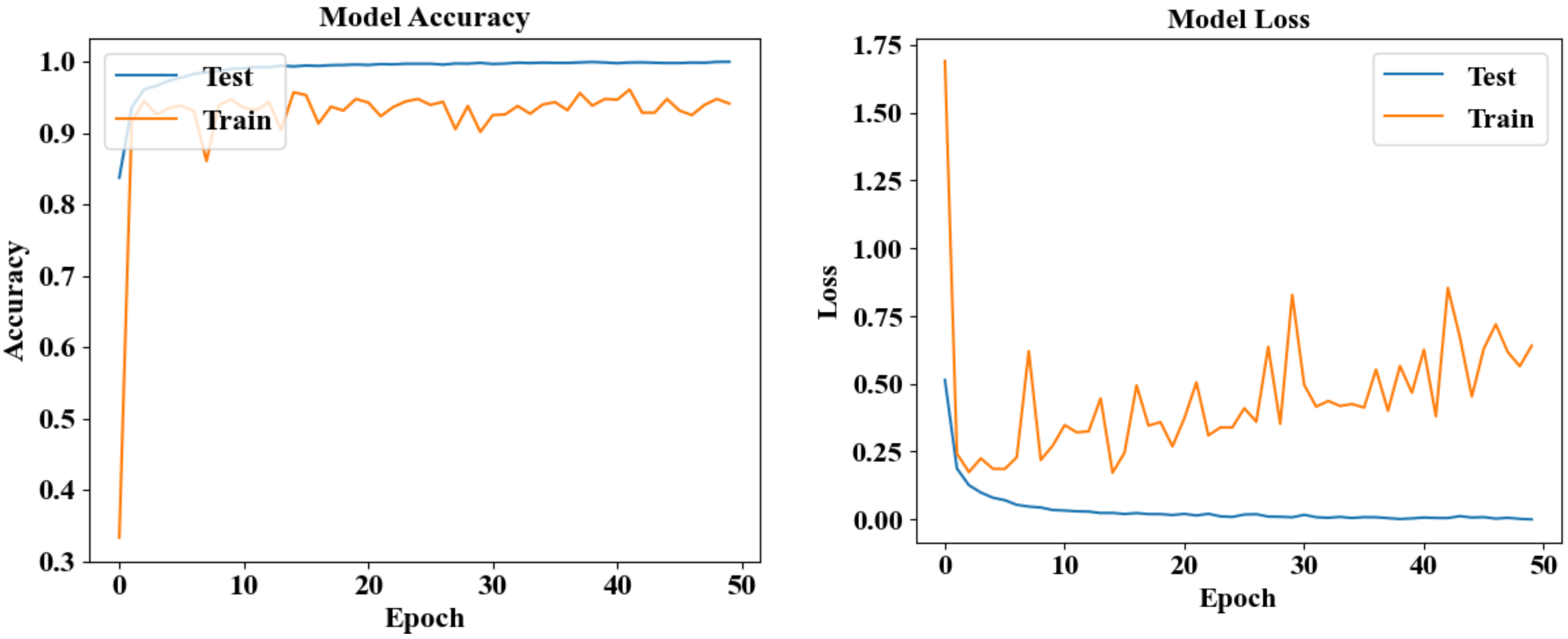

Fig. 4: Accuracy for RMSprop optimizer (i) Loss (ii) Accuracy

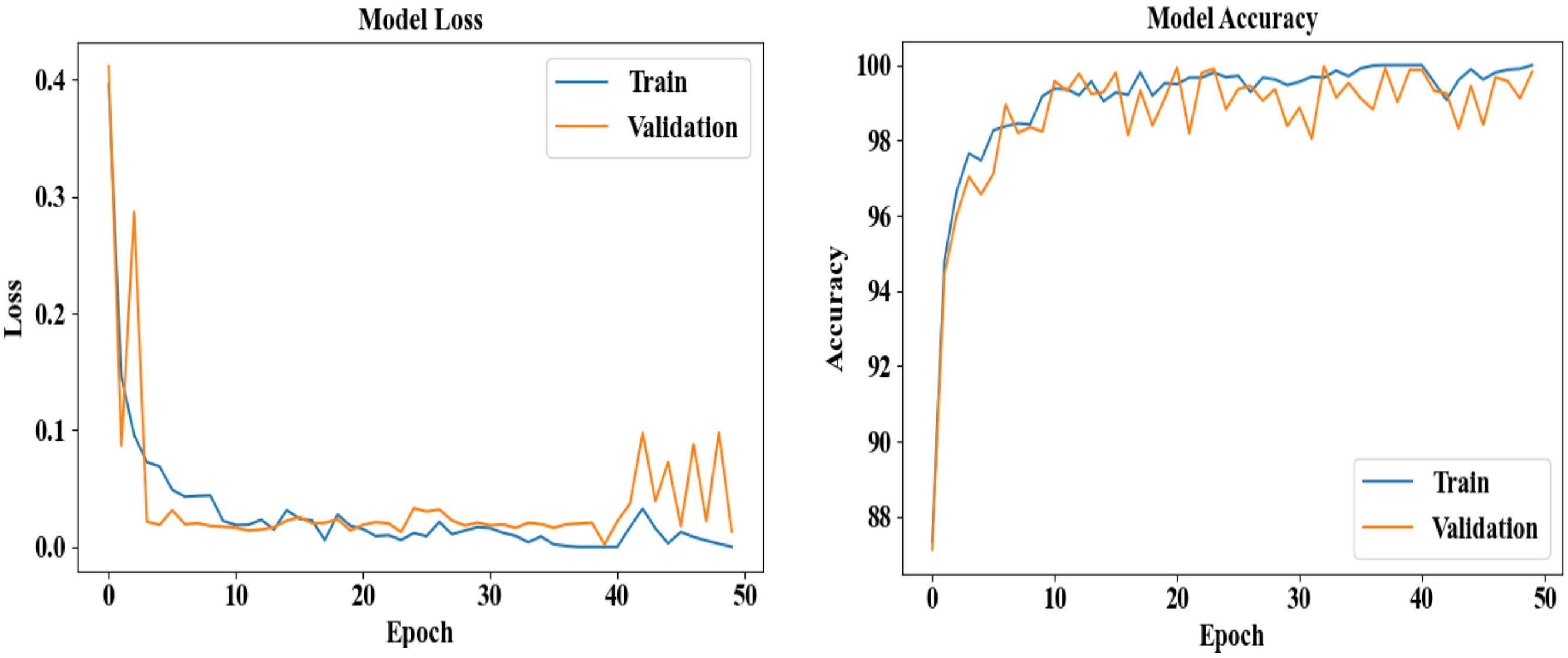

Fig. 5: Accuracy for Adam optimizer (i) Loss (ii) Accuracy

**Table 6:** Performance of DCENWCNet for Raabin-WBC dataset of the RMSprop optimizer

| Class Label | Precision | Recall | F1 Score | Specificity | Support |
|---|---|---|---|---|---|
| Basophil | 1 | 0.9775 | 0.9886 | 1 | 89 |
| Eosinophil | 0.5793 | 0.9751 | 0.7268 | 0.9432 | 322 |
| Lymphocyte | 0.9674 | 0.9497 | 0.9585 | 0.9900 | 1034 |
| Monocyte | 0.7351 | 0.9017 | 0.8099 | 0.9814 | 234 |
| Neutrophil | 0.9946 | 0.9003 | 0.9451 | 0.9922 | 2660 |
| **AV** | **0.8553** | **0.9408** | **0.8858** | **0.9814** | **867.8** |
| **STD** | **0.1894** | **0.0379** | **0.1122** | **0.0223** | |

**Accuracy: 93.93%**

**Table 7:** Performance of DCENWCNet for Raabin-WBC dataset of an Adam optimizer

| Class Label | Precision | Recall | F1 Score | Specificity | Support |
|---|---|---|---|---|---|
| Basophil | 1 | 0.9888 | 0.9944 | 1 | 89 |
| Eosinophil | 0.9717 | 0.9820 | 0.9832 | 0.9791 | 322 |
| Lymphocyte | 0.9625 | 0.9681 | 0.9653 | 0.9882 | 1034 |
| Monocyte | 0.9695 | 0.9846 | 0.9931 | 0.9849 | 234 |
| Neutrophil | 0.9872 | 0.9853 | 0.9810 | 0.9803 | 2660 |
| **AV** | **0.9782** | **0.9817** | **0.9834** | **0.9865** | **867** |
| **STD** | **0.0152** | **0.0080** | **0.0117** | **0.0084** | |

**Accuracy: 98.53%**

### 5.2 Performance metrices on image classification

In this section, we assessed the statistical performance of our method using various evaluation metrics. Specifically, we evaluated recall, precision, and F1-score for each class prediction to measure the method's effectiveness. Statistical examinations of classification performance often indicate the level of certainty in the class prediction made by a DL model. These metrics are shown in Table 6 & Table 7 for optimizer RMSProp & Adam, respectively. An ideal model should have unity values for these parameters, indicating that a higher classification performance is achieved as the value approaches unity. From Tables 6 and 7, we obtain the following key results:

- Given that the accuracy, recall, and F1 score values in Table 6 are consistently close to 0.98, we may anticipate that the proposed models will effectively categorize the WBC classification.

- According to this experimental study, the accuracy obtained is 93.93% with the RMSProp optimizer. Similarly, the performance accuracy for the Adam optimizer is 98.53%. A comparison of overall accuracy rates reveals that the proposed model is quite significant for image classification when by the Adam optimizer.
- For example, of Basophil class (see. Table 6), the DCENWCNet model achieves perfect precision and specificity, with a recall of 0.9888 and an F1 score of 0.9944, implying robust and reliable classification with a support of 89.
- The model shows robust performance in classifying WBC images in the Raabin dataset, with high scores across crucial metrics: an average precision of 0.9782 reflects accurately predicted positive cases; an average recall of 0.9817 confirms the effective identification of true positives; an average F1 score of 0.9834 highlights a well-balanced precision and recall; and an average specificity of 0.9865 shows precise identification of true negatives, underlining the model's overall effectiveness.
- The performance indices indicate that DCENWCNet, with the Adam optimizer, is highly effective for classifying WBCs in the Raabin-WBC dataset. The model indicates excellent precision, recall, F1 scores, and specificity, with consistent performance across all classes, making it a reliable tool for medical diagnostics involving white blood cells.

In order to provide a visual representation of the proposed model's classification capability, Fig. 6 displays the classification outputs of a collection of unseen images (i.e., testing dataset). Each image has a label on top that indicates its actual class which is the image's original label as well as its predicted class. The values for each class (0: Basophils; 1: Eosinophils; 2: Lymphocytes; 3: Monocytes; 4: Neutrophils) are shown in Fig. 6. For accurate image classification, the predicted class (Pre:) must match the actual class (Act:). The following classifications were derived from the given datasets:

- Fig. 6 presents samples from all five classes, where the predicted class labels accurately match the actual class labels for each sample.
- As an example, the top left picture in Fig. 6 belongs to the real class of Basophil (Act: Basophil), and the ensemble architecture prediction for the same image is similarly Basophil (Pre: Basophil). The other images also show instances that are similar.
- A Neutrophils (Act: Neutrophils) class is identified as Lymphocytes (Pre: Lymphocytes) in the fourth column of the second row of the figure, for example, on an infrequent misclassify detection.
- Therefore, the suggested model accurately predicts almost all occurrences of the predicted class, even with the considerable visual changes within the same class.

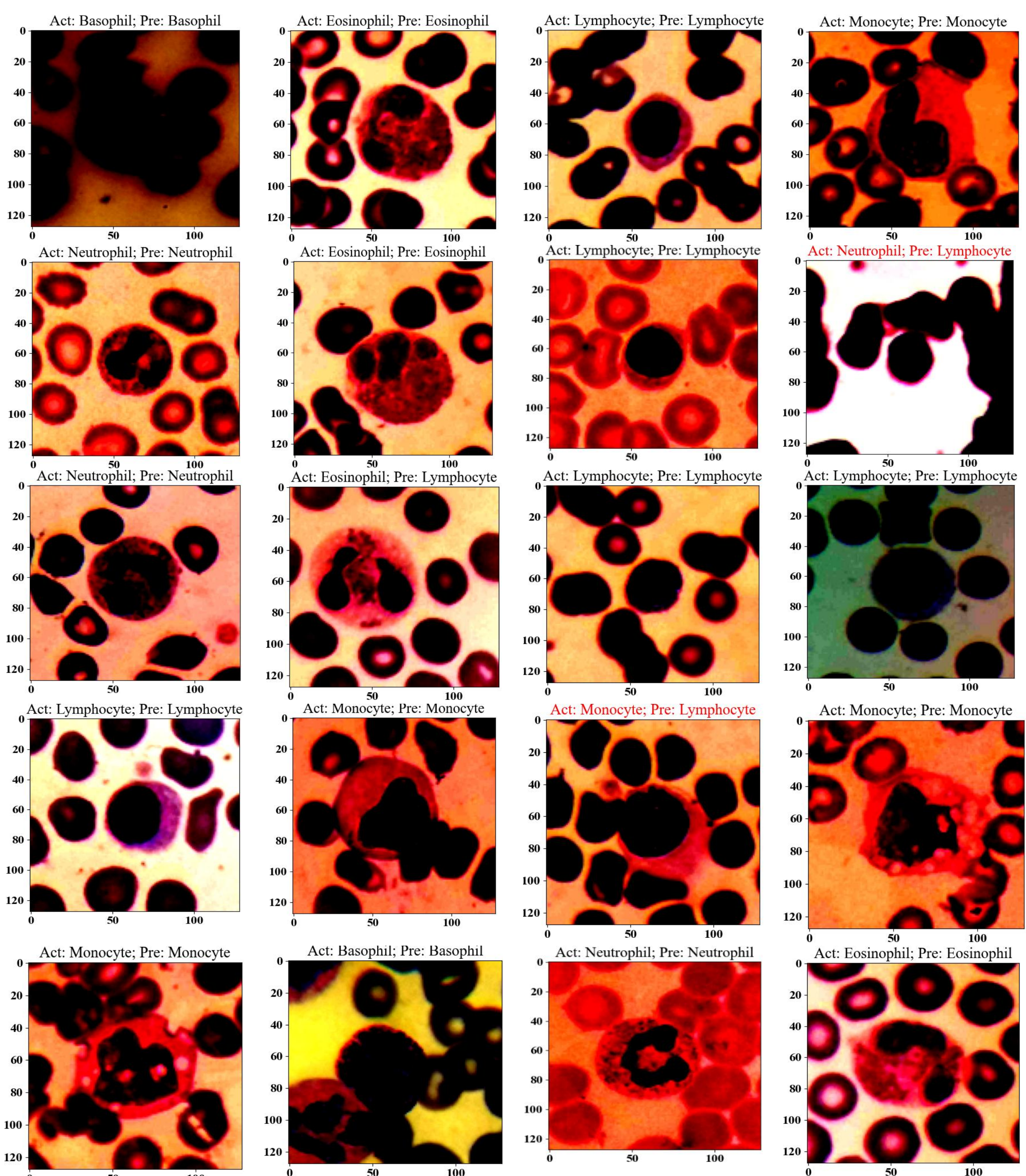

Fig. 6: Example of classified images (Act: actual class; Pre: predicted class, the red colour texts indicate a misclassified instance).

## 5.3 Confusion matrix

A confusion matrix is applied to summarize the output of a classification system. Classification accuracy by itself may be misleading if the dataset has more than two classes or if the number of observations in each class is different. The confusion matrix is made up of four major features (numbers) that define the classifier's metric of measurement. TP, TN, FP, FN are the four numbers. One may have a deeper comprehension of the categorization model's achievements and shortcomings via the computation of a confusion matrix. The diagonal cells should approach the deep learning model to function well, indicating the model's accurate prediction for the relevant class. In both figures, the first confusion matrix is a normalized confusion matrix, where the values represent the proportion of correctly and incorrectly classified instances out of the total number of true instances for each cell type. The second matrix shows raw counts of classified instances, providing insight into the actual number of correct and incorrect predictions.

Fig. 7 (i-ii) and Fig. 8 (i-ii) showcase the confusion matrices for the average and weighted ensemble models recommended, utilizing both the RMSProp and Adam optimizers. The best model attains an accuracy of 98.93% after only 50 epochs for 5 classes. Fig. 8(i) reveals that high values on the diagonal indicate strong model performance, with Basophil and Lymphocyte classifications showing near-perfect accuracy (1.00 and 0.97, respectively), while Monocyte classification is slightly less accurate (0.89). In Fig. 8(ii) presents the DCENWCNet model correctly predicted 89 Basophils, 295 Eosinophils, and 2587 Neutrophils, but misclassified some instances, such as 45 Eosinophils misclassified as Neutrophils. From both Fig. 7 and Fig. 8, the Adam optimizer exhibits superior overall classification accuracy, particularly for Neutrophils and Lymphocytes. Conversely, the RMSprop optimizer scores in fewer misclassifications for Neutrophils and exhibits slightly superior performance for Eosinophils.

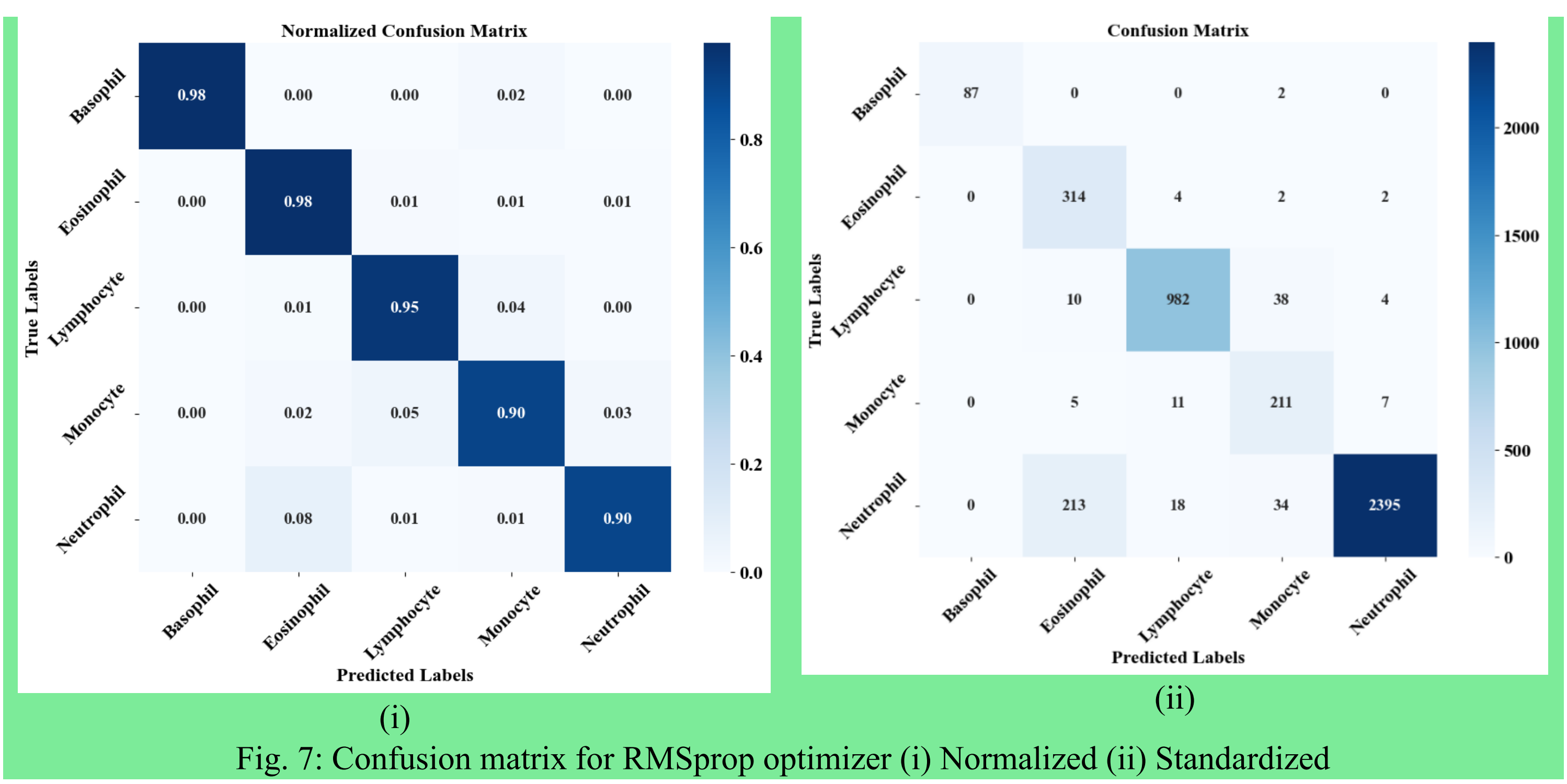

(i)

(ii)

Fig. 7: Confusion matrix for RMSprop optimizer (i) Normalized (ii) Standardized

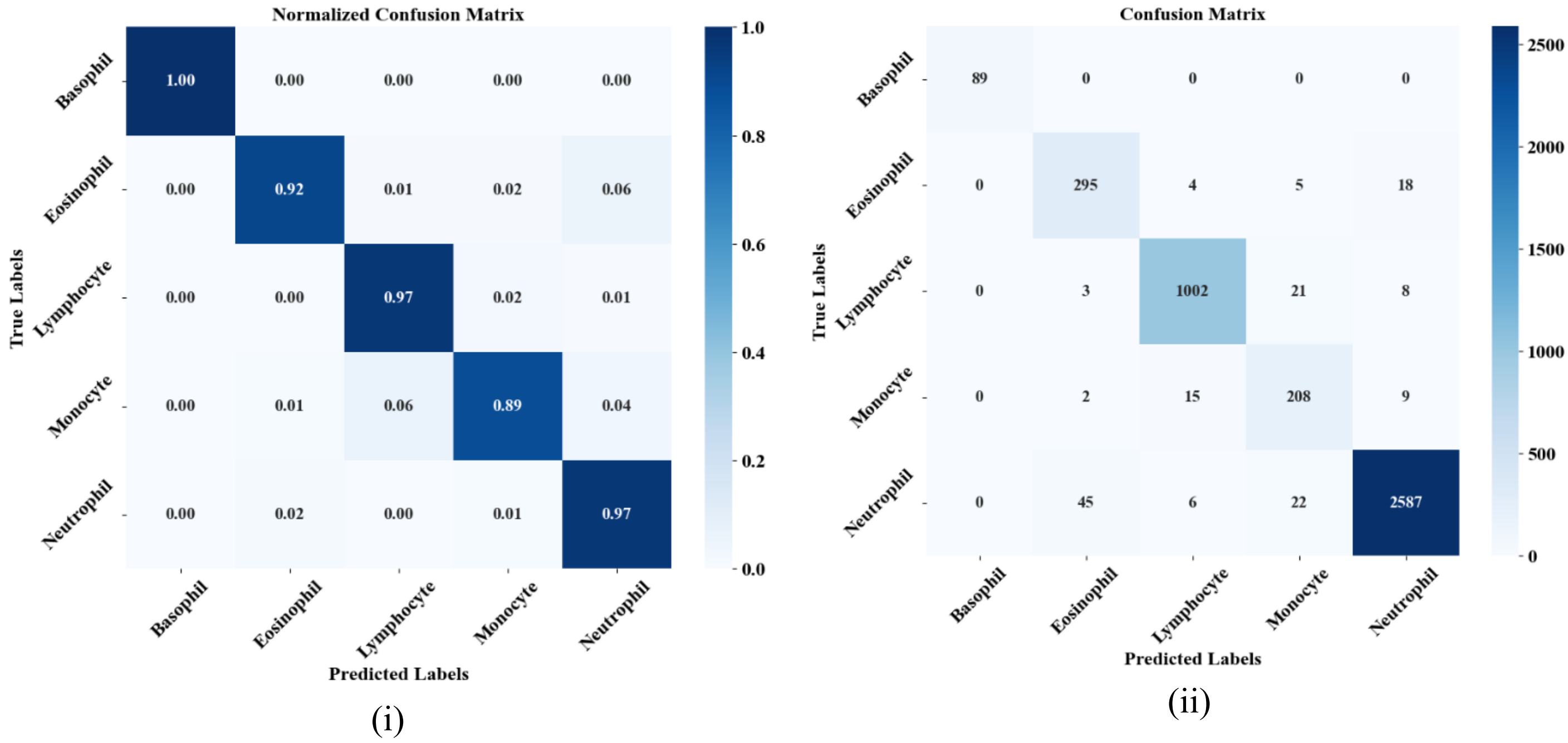

Fig. 8: Confusion matrix for Adam optimizer (i) Normalized (ii) Standardized

### 5.4 ROC with testing

We utilized the ROC curve to assess classifier performance by plotting the true positive rate (TPR = TP/(TP + FN)) against the false positive rate (FPR = FP/(TN + FP)) for the RMSProp and Adam optimizers. To ensure reliable predictions on new data, the dataset was split into 80% for training and 20% for testing using the scikit-learn library, maintaining balanced representation across all five classes. Figs. 9(i) and 9(ii) depict the ROC curves for the proposed DCENWCNet models optimized with RMSProp and Adam, respectively. The ROC curve and AUC score were computed for each class under this evaluation framework. Efficient algorithms, transfer learning strategies, and optimization techniques such as RMSProp and Adam contribute significantly to reducing model training time. Figures 9(i) and 9(ii) illustrate the multi-class ROC curves corresponding to the DCENWCNet models optimized with RMSProp and Adam, respectively. For each optimizer, class-wise ROC curves and the associated AUC values are computed using a one-vs-rest evaluation strategy, along with micro- and macro-averaged ROC curves to provide an overall performance assessment. As observed in Figs. 9(i) and 9(ii), the proposed model achieves consistently high AUC values across all five WBC classes, with mean AUC values close to 0.99 for both optimizers. These results indicate strong discriminative capability and stable generalization performance on the test set. Furthermore, the use of efficient optimization techniques such as RMSProp and Adam contributes to effective convergence and robust classification performance without compromising predictive accuracy.

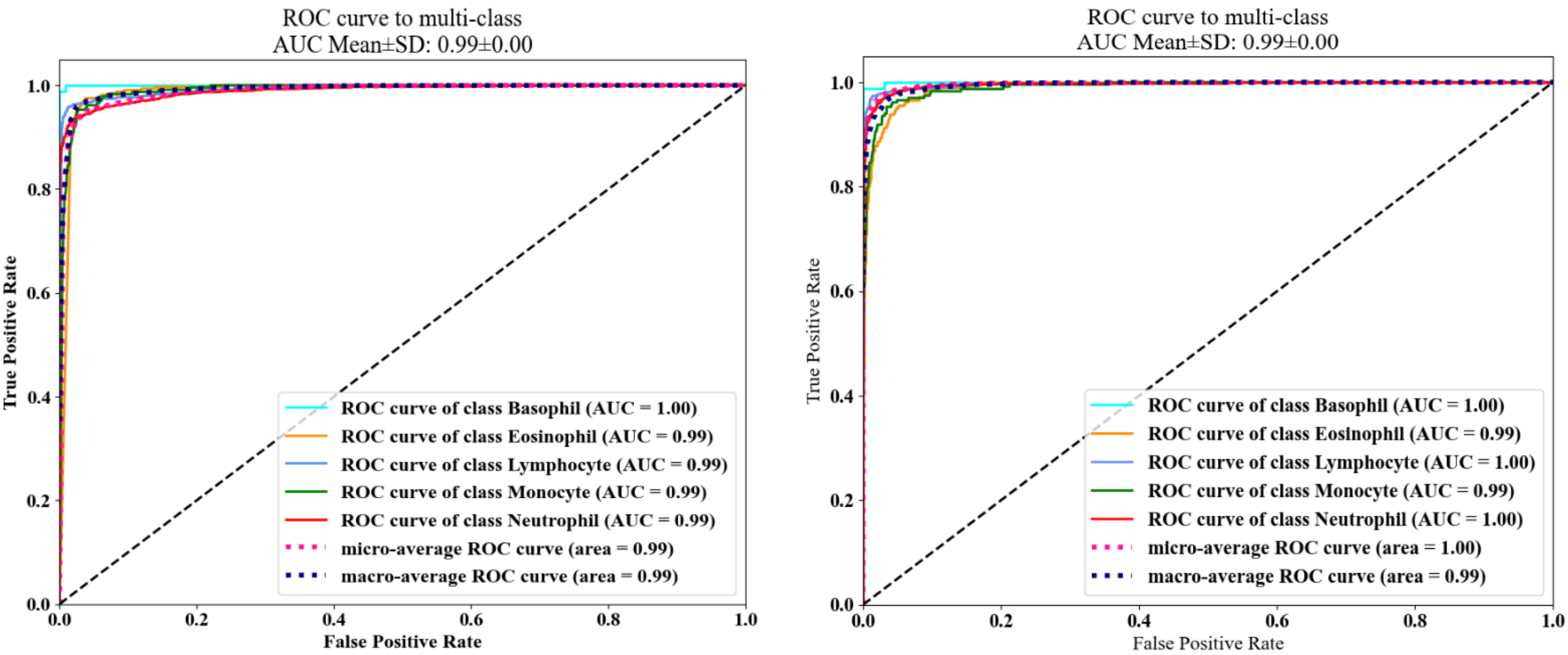

Fig. 9: Multi-class ROC curves and corresponding AUC scores for the proposed DCENWCNet model optimized using (i) RMSProp and (ii) Adam.

### 5.5 LIME explainable

The images Fig. 10 illustrate an original cell image and its corresponding LIME (Local Interpretable Model-agnostic Explanations) segmentation. LIME is employed in the proposed framework to highlight how established explainability techniques, when integrated with carefully designed ensemble architectures, can provide reliable and clinically interpretable decision support without incurring excessive computational overhead. The first image presents a microscopic view of a cell stained in dark purple, surrounded by other cells or cellular debris, labeled as "Basophil, Eosinophil, Lymphocyte, Monocyte, and Neutrophil." The all image demonstrates LIME segmentation for the same cell, highlighting specific regions in green identified by the LIME algorithm as significant for classifying the cell. The highlighted areas on the cell and its surroundings suggest these regions contain features crucial for the model's classification. This comparative visualization provides insight into both the cell's appearance and the interpretability of the classification model, emphasizing the role of explainable AI in medical imaging and diagnosis.

The LIME visualizations consistently highlight diagnostically relevant regions across all white blood cell classes, primarily focusing on the nucleus shape, granularity, and cytoplasmic texture while largely suppressing background WBC. High confidence scores (≈0.85–1.00) indicate that the model bases its predictions on clinically meaningful morphological cues rather than spurious artifacts. The class-specific highlighted patterns such as lobulated nuclei in neutrophils, dense granules in basophils and eosinophils, and large nuclei in lymphocytes demonstrate that the proposed model learns interpretable and pathology-aligned features, supporting its reliability for clinical decision support.

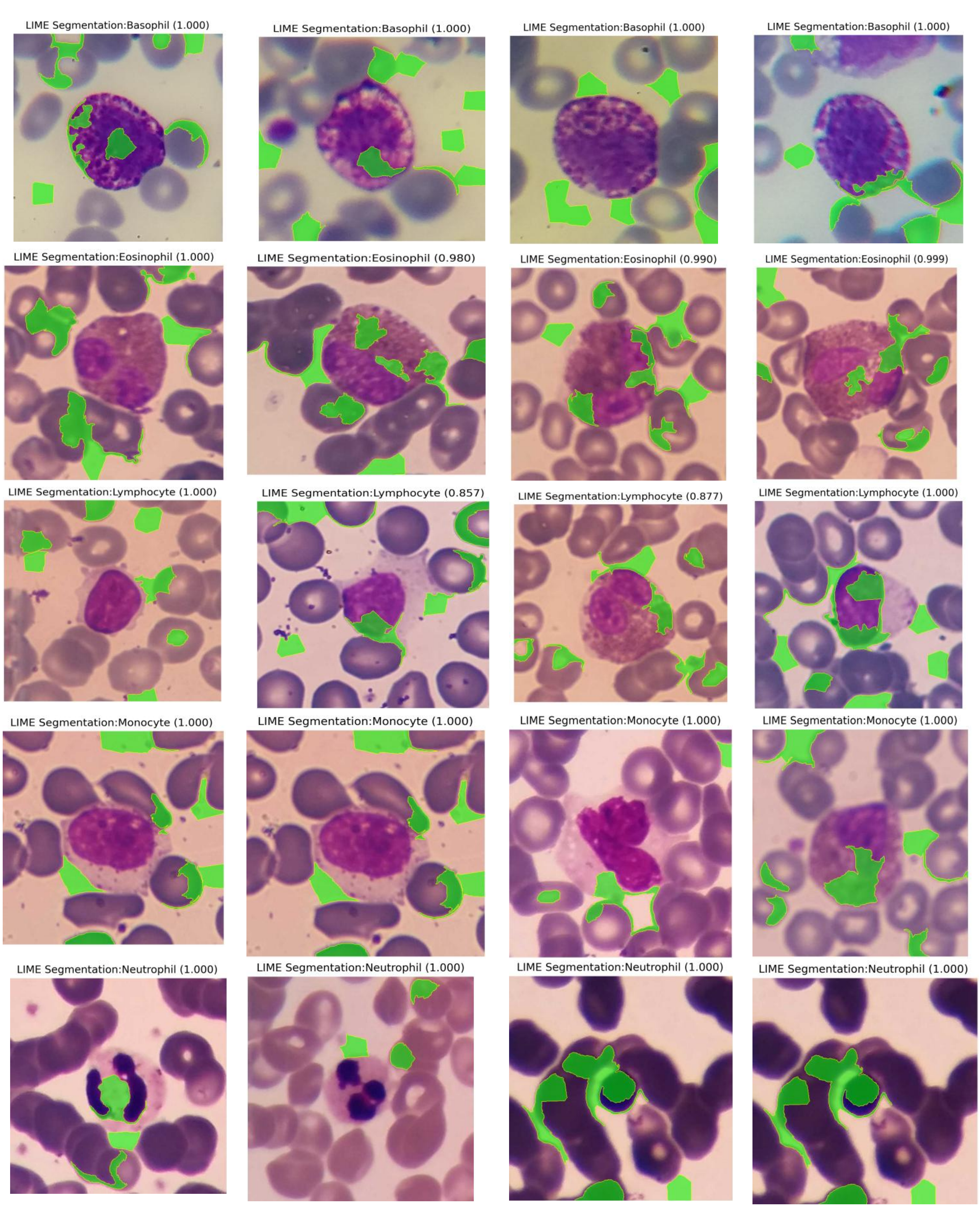

Fig. 10: Highlighting significant features for classification through LIME segementation

## 5.6 Comparative performance analysis

This section presents a comprehensive comparative performance analysis of the proposed model against existing state-of-the-art methods. The evaluation includes quantitative performance comparisons, statistical significance analysis, and comparisons with prior studies to highlight the effectiveness and robustness of the proposed approach. Statistical tests are employed to verify whether the observed performance improvements are statistically significant rather than incidental, thereby providing strong empirical evidence of the superiority of the DCENWCNet model over competing techniques.

### *I. Compare with existing pre-trained models*

Table 8 summarizes the mean values of the evaluation metrics obtained across 30 independent runs for all 12 deep learning models, where 80% of the available data was used for training. In general, architectures that are deeper, more intricate, and contain a larger number of trainable parameters tend to require longer training and inference times, which directly affects computational cost. When the proposed DCENWCNet (Adam) is compared with the benchmark pre-trained networks, clear and practically meaningful performance gaps are observed. Table 8 provides a side-by-side comparison between the proposed approach and the competing models, and the resulting improvements are supported by multiple statistical indicators, as discussed in the following points:

- The proposed model (Adam) has a precision of 0.9782, which is markedly higher compared to the next best, AlexNet, at 0.9428. Precision indicates the accuracy of positive predictions, and the proposed model suggests far fewer false positives than any other model listed in Table 7.
- Relative to VGG16 and VGG19, the proposed DCENWCNet (Adam) delivers clear improvements in predictive performance. Specifically, it increases accuracy by 7.94% over VGG16 and by 6.07% over VGG19, while also showing corresponding gains in precision, recall, and F1-score. Notably, these improvements are achieved with a noticeably shorter training time, indicating that the proposed model is not only more accurate but also more computationally efficient.
- With a recall of 0.9817, the proposed model again ranks first among all methods, with DenseNet being the closest competitor at 0.9707. Since recall reflects how effectively a classifier captures all relevant instances, this result indicates that DCENWCNet identifies true positives more reliably.
- The proposed approach achieves an F1-score of 0.9834, which is notably higher than the next best result reported for DenseNet (0.9542). Because the F1-score combines precision and recall through their harmonic mean, this outcome suggests a stronger overall balance between missed detections and false alarms.
- The accuracy of DCENWCNet reaches 98.53%, exceeding the second-best model (DenseNet: 87.12%) by a wide margin. This metric reflects the overall proportion of correctly classified samples across all categories.
- Compared with the classical SVM baseline, DCENWCNet (Adam) increases accuracy from 96.04% to 98.53%, while also improving precision, recall, and F1-score. This comparison highlights the benefit of the deep ensemble design over a traditional machine-learning classifier.

- Against lightweight architectures such as MobileNetV2 and EfficientNetV2, the proposed model achieves higher accuracy (98.53% vs. 95.33% and 97.77%, respectively) and maintains consistently strong performance across all reported evaluation measures, while keeping training cost competitive.
- In addition to these performance gains, DCENWCNet (Adam) remains computationally efficient, requiring approximately 40 minutes of training time. This runtime is shorter than that of the other pre-trained networks, which still report inferior scores across the evaluated metrics relative to the proposed method.

**Table 8:** Performance metrics of various pre-trained methods applied to augmented WBC data

| **Method** | **Precision** | **Recall** | **F1-score** | **Acc. (%)** | **Training time** |
|---|---|---|---|---|---|
| RCNN | 0.9136 | 0.9256 | 0.9125 | 93.25 | 1h 10min 16s |
| AlexNet | 0.9428 | 0.9436 | 0.9439 | 93.45 | 1h 29min 35s |
| VGG16 | 0.7831 | 0.7429 | 0.6850 | 90.59 | 1h 48min 28s |
| VGG19 | 0.8806 | 0.9194 | 0.7700 | 92.46 | 1h 62min 11s |
| SqueezeNet-FCM | 0.9325 | 0.9433 | 0.9136 | 94.28 | 1h 54min 40s |
| Inception v3 | 0.7839 | 0.8831 | 0.8247 | 89.56 | 58min 20s |
| DenseNet | 0.9402 | 0.9707 | 0.9542 | 97.12 | 52min 42s |
| ResNet50 | 0.9287 | 0.9615 | 0.9428 | 96.36 | 48min 52s |
| SVM | 0.9723 | | 0.9614 | 96.04 | 1hr 52min 10s |
| MobileNetV2 | 0.8651 | 0.9478 | 0.8950 | 95.33 | 49min 39s |
| EfficientNetv2 | 0.9714 | 0.9841 | 0.9821 | 97.77 | 48min 35s |
| **DCENWCNet (RMSProp)** | **0.8553** | **0.9408** | **0.8858** | **93.53** | **42min 04s** |
| **DCENWCNet (Adam)** | **0.9782** | **0.9817** | **0.9834** | **98.53** | **40min 04s** |

In our analysis, the DCENWCNet (Adam) model demonstrated the shortest execution time during training and exhibited the quickest system response in test evaluations. As shown in Table 9, the proposed DCENWCNet introduces a moderate increase in computational complexity due to its ensemble structure. However, this increase is well controlled, as DCENWCNet (Adam) achieves a low inference time of 5 ms, with only 1.10 G FLOPs and 6.8 M parameters. Compared with heavyweight models such as VGG16 and RCNN, the proposed model significantly reduces computational cost, while maintaining near real-time performance. Although slightly more complex than single lightweight networks, DCENWCNet provides a favorable trade-off between efficiency and classification accuracy, making it suitable for practical clinical deployment. Although some lightweight models show similar latency, they exhibit inferior classification performance. Overall, DCENWCNet offers a favorable trade-off between high diagnostic accuracy and efficient inference, enabling reliable detection and identification of WBCs for automated hematological analysis. Although the quantitative improvements over recent state-of-the-art methods are significant, such gains are clinically significant in hematological image analysis, where even small improvements in accuracy and recall can reduce misclassification risks.

**Table 9:** Inference efficiency and computational complexity comparison.
**"–" indicates that FLOPs and parameter counts are not reported for SVM (Cortes & Vapnik, 1995; Schölkopf & Smola, 2002)**

| Method | Inference time (ms) | FLOPs (G) | Parameters (M) |
|---|---|---|---|
| RCNN | 14 | 15.2 | 134.0 |
| AlexNet | 11 | 0.72 | 61.0 |
| VGG16 | 35 | 15.5 | 138.3 |
| VGG19 | 28 | 13.6 | 139.6 |
| ResNet50 | 28 | 4.1 | 25.6 |
| SqueezeNet-FCM | 17 | 0.83 | 1.5 |
| Inception-v3 | 39 | 5.7 | 23.8 |
| DenseNet | 14 | 2.9 | 8.0 |
| SVM | 8 | - | - |
| MobileNetV2 | 11 | 0.30 | 3.4 |
| EfficientNetV2 | 8 | 0.43 | 21.5 |
| **DCENWCNet (RMSProp)** | **12** | **1.10** | **6.8** |
| **DCENWCNet (Adam)** | **5** | **1.10** | **6.8** |

***II. Compare with statistical result***

As shown in Table 10, using the Raabin-WBC dataset, the proposed DCENWCNet (Group B) exhibits clear percentage-wise performance gains over pre-trained models (Group A), together with statistical significance. In terms of Precision, DCENWCNet achieves 97.82%, representing an improvement of approximately +0.6% to +19.5% over most baselines such as VGG16, Inception v3, MobileNetV2, and ResNet50, with several comparisons reaching statistical significance ($p \leq 0.05$–$0.001$). For Recall, DCENWCNet attains 98.17%, corresponding to gains of about +1.4% to +23.8%, particularly significant when compared with VGG16, DenseNet, and ResNet50 ($p \leq 0.01$ and $p \leq 0.001$). The most notable improvement is observed in the F1-score, where DCENWCNet reaches 98.34%, yielding an increase of roughly +2.1% to +43.6% over baseline models, with multiple statistically significant differences ($p \leq 0.05$, $p \leq 0.01$, and $p \leq 0.001$), indicating a substantially better balance between precision and recall. Finally, in terms of Accuracy, DCENWCNet achieves 98.53%, improving upon competing models by approximately +0.8% to +9.0%, with statistically significant gains against several strong baselines such as VGG16, DenseNet, SVM, MobileNetV2, and EfficientNetV2. Overall, these percentage improvements, together with consistent statistical significance, confirm that DCENWCNet delivers not only numerically higher performance but also statistically reliable and meaningful improvements on the Raabin-WBC dataset.

Table 10. Comparison of pre-trained models (Group A) and the proposed model DCENWCNet (Group B) metrics, *p≤0.05, **p≤0.01, ***p≤0.001

| Matrices | Pre-trained model | Mean ± Standard deviation | | t-statistic | p-value |
|---|---|---|---|---|---|
| | | Group A | Group B (DCENWCNet) | | |
| Precision | RCNN | 0.9136 | | -0.725 | 0.325 |
| | AlexNet | 0.9428 | | -0.617 | 0.219 |
| | VGG16 | 0.7831 | | -4.913 | 0.497 |
| | VGG19 | 0.8806 | | -1.623 | 0.000*** |
| | SqueezeNet-FCM | 0.9325 | | -0.479 | 0.623 |
| | Inception v3 | 0.7839 | 0.9782 ± 0.0152 | -6.145 | 0.186 |
| | DenseNet | 0.9402 | | -0.201 | 0.368 |
| | ResNet50 | 0.9287 | | -0.171 | 0.412 |
| | SVM | 0.9723 | | -0.973 | 0.051* |
| | MobileNetV2 | 0.8651 | | -1.285 | 0.654 |
| | EfficientNetv2 | 0.9714 | | -0.917 | 0.005* |
| Recall | RCNN | 0.9256 | | -3.108 | 0.481 |
| | AlexNet | 0.9436 | | -2.198 | 0.034 |
| | VGG16 | 0.7429 | | -8.947 | 0.005** |
| | VGG19 | 0.9194 | | -0.2188 | 0.45 |
| | SqueezeNet-FCM | 0.9433 | | -3.371 | 0.921 |
| | Inception v3 | 0.8831 | 0.9817 ± 0.0080 | -10.275 | 0.532 |
| | DenseNet | 0.9707 | | -2.103 | 0.012* |
| | ResNet50 | 0.9615 | | -2.331 | 0.000*** |
| | SVM | | | | |
| | MobileNetV2 | 0.9478 | | -0.519 | 0.201 |
| | EfficientNetv2 | 0.9841 | | -0.101 | 0.028 |
| F1 score | RCNN | 0.9125 | | -4.114 | 0.267 |
| | AlexNet | 0.9439 | | -2.251 | 0.046* |
| | VGG16 | 0.685 | | -9.128 | 0.574 |
| | VGG19 | 0.77 | | -12.671 | 0.653 |
| | SqueezeNet-FCM | 0.9136 | | -6.146 | 0.006** |
| | Inception v3 | 0.8247 | 0.9834 ± 0.0117 | -11.328 | 0.418 |
| | DenseNet | 0.9542 | | -4.452 | 0.000*** |
| | ResNet50 | 0.9428 | | -2.182 | 0.145 |
| | SVM | 0.9614 | | -1.271 | 0.046* |
| | MobileNetV2 | 0.895 | | -6.178 | 0.601 |
| | EfficientNetv2 | 0.9821 | | -0.461 | 0.002** |
| Acc. (%) | RCNN | 93.25 | | -4.161 | 0.367 |
| | AlexNet | 93.45 | 98.53 ± 0.0029 | -4.002 | 0.308 |
| | VGG16 | 90.59 | | -8.956 | 0.000*** |

| | | | |
|---|---|---|---|
| VGG19 | 92.46 | -3.089 | 0.028* |
| SqueezeNet-FCM | 94.28 | -2.991 | 0.784 |
| Inception v3 | 89.56 | -13.459 | 0.661 |
| DenseNet | 97.12 | -0.45 | 0.034** |
| ResNet50 | 96.36 | -0.985 | 0.926 |
| SVM | 96.04 | -1.273 | 0.005** |
| MobileNetV2 | 95.33 | -2.467 | 0.029* |
| EfficientNetv2 | 97.77 | -0.328 | 0.038* |

### ***III. Compare with existing literature***

Table 11 presents a comparison with previously reported results on the Raabin-WBC dataset. Building on recent advances in the literature, our study combines a strong augmentation pipeline with a pre-trained deep learning backbone to improve recognition performance. As shown in Table 8, the proposed method achieves 98.53% accuracy, with precision = 0.9834, recall = 0.9817, F1-score = 0.9782, and specificity = 0.9865. Collectively, these outcomes indicate that pairing effective augmentation with established pre-trained architectures can substantially strengthen WBC classification. The following points summarize how the proposed model compares with earlier studies:

- Sharma et al. (2019) projected a deep CNN for early WBC detection, achieving an accuracy 5% lower than that of our proposed method.
- Wei et al. (2020) proposed a lightweight WBC recognition approach that combines ResNet/DenseNet feature learning with a spatial–channel attention module (SCAM). When contrasted with DCENWCNet, their reported results indicate lower performance, with approximate gaps of 0.18 in accuracy, 0.59 in precision, 0.35 in recall, and 0.56 in F1-score in favor of the proposed method.
- Tavakoli et al. (2021) introduced a hybrid strategy that integrates pre-trained models with SVM/CNN components. Relative to their outcomes, DCENWCNet yields an approximate 4.1% gain in accuracy, together with improvements of about 0.6% in precision, 3.2% in recall, and 2.3% in F1-score. In addition, it offers a clear computational benefit by reducing training time from ~1 h 52 min to ~40 min, which corresponds to roughly a 64% reduction in training duration, supporting its practicality for time-constrained clinical settings.
- Jiang et al. (2022) employed multiple advanced CNN architectures (including DRFA-Net) to better localize WBC regions and improve downstream classification. Despite these enhancements, the reported performance remains below that achieved by DCENWCNet (Adam) in our experiments.
- Han et al. (2023) developed an improved framework using an enhanced residual convolution block, hybrid spatial pyramid pooling, coordinate attention, and the efficient intersection over union (EIOU) strategy. While their modifications strengthen WBC classification, the proposed DCENWCNet built in the same spirit of leveraging modern architectural refinements—still delivers a modest additional improvement across the main evaluation measures.

- Compared to the approach of Li and Liu (2023), the proposed model DCENWCNet optimized with Adam demonstrates an approximate performance relative enhancement of 0.75%, along with more consistent and statistically significant gains across evaluation metrics, indicating enhanced robustness and discriminative capability.
- Rubin et al. (2023) established a weighted ensemble Deep Vision Transformer (ViT). While they observed differences of approximately 4% in accuracy, F1 score, recall, and precision compared to our research work, their specificity results were better.
- Relative to latest studies by Erten et al. (2024) and Li et al. (2025), the proposed model DCENWCNet demonstrates consistent relative performance improvements across all evaluation metrics. In particular, DCENWCNet achieves relative gains of approximately 0.78%–0.80% in accuracy, 0.05%–2.7% in precision, 0.35%–0.46% in recall, 0.77%–2.4% in specificity, and 0.54%–2.1% in F1-score, indicating more balanced and robust classification performance. Moreover, these enhancements are obtained with substantially reduced training time (40 min 04 s), compared to over one hour reported in recent studies (Li et al. (2025), highlighting the improved computational efficiency of the proposed framework.

Finally, the proposed method achieves the highest accuracy and exhibits consistently well-balanced results across the remaining evaluation measures. This consistency indicates strong robustness and reliable classification capability when compared with the alternatives summarized in Table 11.

**Table 11:** Benchmark Table under the existing literature support

| **Author** | **Acc.%** | **Precision** | **Recall** | **Specificity** | **F1** | **Training time** |
|---|---|---|---|---|---|---|
| Sharma et al., (2019) | 95.99 | 0.82.62 | 0.9508 | × | 0.9350 | × |
| Tavakoli et al., (2021) | 94.65 | 0.9723 | 0.9507 | × | 0.9614 | 1hr 52min 10s |
| Jiang et al., (2022) | 95.17 | 0.9043 | 0.9340 | × | 0.9189 | × |
| Chen et al., (2022) | 98.71 | 0.9718 | 0.9842 | × | 0.9778 | × |
| Li and Liu (2023) | 97.8 | × | × | × | × | × |
| Han et al., (2023) | 93.97 | × | × | × | 0.9703 | × |
| Rivas-Posada & Chacon-Murguia, (2023) | 97.69 | 0.9565 | - | - | 0.9677 | 58min 25s |
| Rubin et al., (2023) | 97 | 0.97 | 0.97 | 0.97 | 0.9700 | × |
| Erten et al. (2024) | 97.77 | 0.9776 | 0.9783 | × | 0.9779 | × |
| Li et al. (2025) | 97.76 | 95.34 | 97.29 | × | 0.9631 | 1hr 7min 2s |
| **Proposed Model (DCENWCNet)** | **98.53** | **0.9782** | **0.9817** | **0.9865** | **0.9834** | **40min 04s** |

**Abbreviation:** '×' not considered the result

## 6. Conclusions

This research work introduces DCENWCNet, a structured ensemble-based CNN framework for accurate and computationally efficient WBC classification. By integrating heterogeneous dropout-based regularization, pixel-level standardization, and class-aware data augmentation across multiple CNN branches, the proposed research work achieves robust feature learning and strong generalization capability. Comprehensive evaluation on the Raabin-WBC dataset demonstrates that DCENWCNet reliably classifies basophils, eosinophils, lymphocytes, monocytes, and neutrophils with consistently high performance, outperforming classical machine learning approaches and widely applied existing pre-trained DL models while maintaining favorable computational efficiency. In addition to strong predictive performance, the proposed model DCENWCNet framework incorporates post-hoc interpretability through LIME to enhance transparency and clinical relevance. The LIME-based visual explanations consistently highlight diagnostically meaningful regions of white blood cells, such as nucleus shape, cytoplasmic texture, and cell boundaries, that are known to be critical for hematological assessment. This confirms that the model's decisions are driven by relevant morphological features rather than spurious background patterns. By providing intuitive and class-specific visual justifications for model predictions, the LIME analysis supports trust, interpretability, and practical adoption of DCENWCNet in computer-aided hematological diagnosis and clinical decision-support systems.

By integrating three deep CNN architectures, the approach improves detection accuracy while reducing computational complexity. Utilizing an augmented dataset, we achieve high accuracy through a balance of network depth and width. Experimental results demonstrate that DCENWCNet achieves a classification accuracy of 98.53%, with consistently high precision, recall, F1-score, and AUC values across all classes. Comparative analysis against classical machine learning models and widely adopted pre-trained deep learning architectures confirms the superiority of the proposed method in terms of both predictive performance and training efficiency. Statistical significance testing further supports the robustness of the observed improvements. These findings indicate that the ensemble regularization strategy, achieved through controlled architectural diversity and probability-based aggregation, plays a critical role in improving discriminative capability while mitigating overfitting and class imbalance effects.From a practical perspective, the proposed framework offers a reliable solution for computer-aided WBC detection and classification, with reduced computational overhead compared to deeper single-network models. This balance between accuracy and efficiency enhances its potential applicability in automated hematological analysis and decision-support systems, particularly in resource-constrained clinical environments.

While the proposed DCENWCNet framework demonstrates strong performance on single-cell WBC images from the Raabin-WBC dataset, certain limitations remain. These include reliance on a single publicly available dataset, evaluation on isolated cell images rather than complex multi-cell blood smears, and the computational overhead inherent to ensemble-based architectures. Future work will extend the proposed DCENWCNet framework to more realistic and challenging scenarios, including multi-cell images containing overlapping and partially occluded white blood cells, which better reflect real blood smear conditions. In addition, the model will be evaluated on additional publicly available wellknon like (LISC, PBC, BCCD) WBC datasets collected under diverse imaging protocols, staining conditions, and acquisition devices, as well as on multi-center clinical datasets. Such cross-dataset validation will enable a more comprehensive assessment of the robustness, generalizability, and domain adaptability of the proposed approach in real-world clinical workflows. Further research will also investigate computationally efficient ensemble variants and adaptive regularization strategies to reduce inference overhead while preserving classification performance. Although larger ensemble sizes and alternative dropout (e.eg., 2, 4, and 5) configurations may influence performance, a systematic ablation analysis of these factors is left for future investigation beyond the scope of the present study. In future work, the proposed framework will be further optimized to enhance computational efficiency by exploring lightweight feature extraction strategies and integrating recent, efficient explainability techniques that can provide rapid and reliable visual interpretations without compromising predictive performance.s

**Acknowledgements**

The authors thank the Editor-in-Chief and the three anonymous reviewers for their esteemed suggestions, which helped enhance the quality of this manuscript. The authors would also like to acknowledge Amrita School of Artificial Intelligence, Amrita Vishwa Vidyapeetham, Bengaluru, the Technical University of Munich, Germany, and the Raabin-WBC dataset contributors for providing their platforms and valuable datasets.

**Competing interests**

All authors declare that they have no known conflicts of interest.

**Ethical Statement**

Not applicable.

**Data Availability**

The datasets used in this study are all public and freely available. Codes will be made available on request.

**Funding**

This research did not receive any specific grant from funding agencies in the public, commercial, or non-profit sectors.